\documentclass{article}

\usepackage[preprint]{neurips_2023}

\usepackage[utf8]{inputenc}
\usepackage{CJKutf8}
\usepackage{amsmath,stmaryrd}   
\usepackage{soul}
\usepackage{arydshln}
\usepackage{xspace}
\usepackage{float}
\usepackage{todonotes}
\usepackage{xcolor} 
\usepackage{multirow} 
\usepackage{colortbl} 
\usepackage{graphicx} 
\usepackage{placeins}
\usepackage{subcaption}
\DefineNamedColor{named}{PineGreen}     {cmyk}{0.92,0,0.59,0.25}
\definecolor{LightGray}{gray}{0.9}
\definecolor{DarkGray}{gray}{0.7}
\usepackage{subcaption}
\usepackage[loose]{minitoc}

\usepackage[utf8]{inputenc} 
\usepackage[T1]{fontenc}    
\usepackage{hyperref}       
\hypersetup{
    colorlinks=true,
    linkcolor=violet,
    citecolor=violet,
    urlcolor=blue
}
\usepackage{url}            
\usepackage{booktabs}       
\usepackage{amsfonts}       
\usepackage{nicefrac}       
\usepackage{microtype}      
\usepackage{xcolor}         
\usepackage{fontawesome}

\title{\textsc{Astraios}: Parameter-Efficient Instruction Tuning Code Large Language Models}

\author{
\vspace{2.5mm} 
\hspace{-6.5mm}
        Terry Yue Zhuo~$^{1,2}$\hspace{6.5mm}
        Armel Zebaze~$^3$\thanks{The work was partially done at Hugging Face.}\hspace{5.5mm}
        Nitchakarn Suppattarachai~$^1$
    \\
    \vspace{2.5mm}
    \textbf{\hspace{-5.5mm}
        Leandro von Werra~$^3$\hspace{4.5mm}
        Harm de Vries~$^4$\hspace{4.5mm}
        Qian Liu~$^5$\hspace{4.5mm}
        Niklas Muennighoff~$^6$
    }
    \\
    $^1$ Monash University\hspace{5mm}
    $^2$ CSIRO's Data61\hspace{5mm}
    $^3$ Hugging Face\\
    $^4$ ServiceNow Research\hspace{5mm}
    $^5$ Sea AI Lab\hspace{5mm}
    $^6$ Contextual AI
    \\
    {\tt \href{mailto:terry.zhuo@monash.edu}{\textcolor{black}{terry.zhuo@monash.edu}}}
    \\[\smallskipamount]
    \faGithub~\url{https://github.com/bigcode-project/astraios}
}

\begin{document}

\maketitle
\begin{tikzpicture}[remember picture,overlay,shift={(current page.north west)}]
\node[anchor=north west,xshift=18.1cm,yshift=-2.7cm]{\scalebox{0.46}[0.45]{\includegraphics[width=5.5cm]{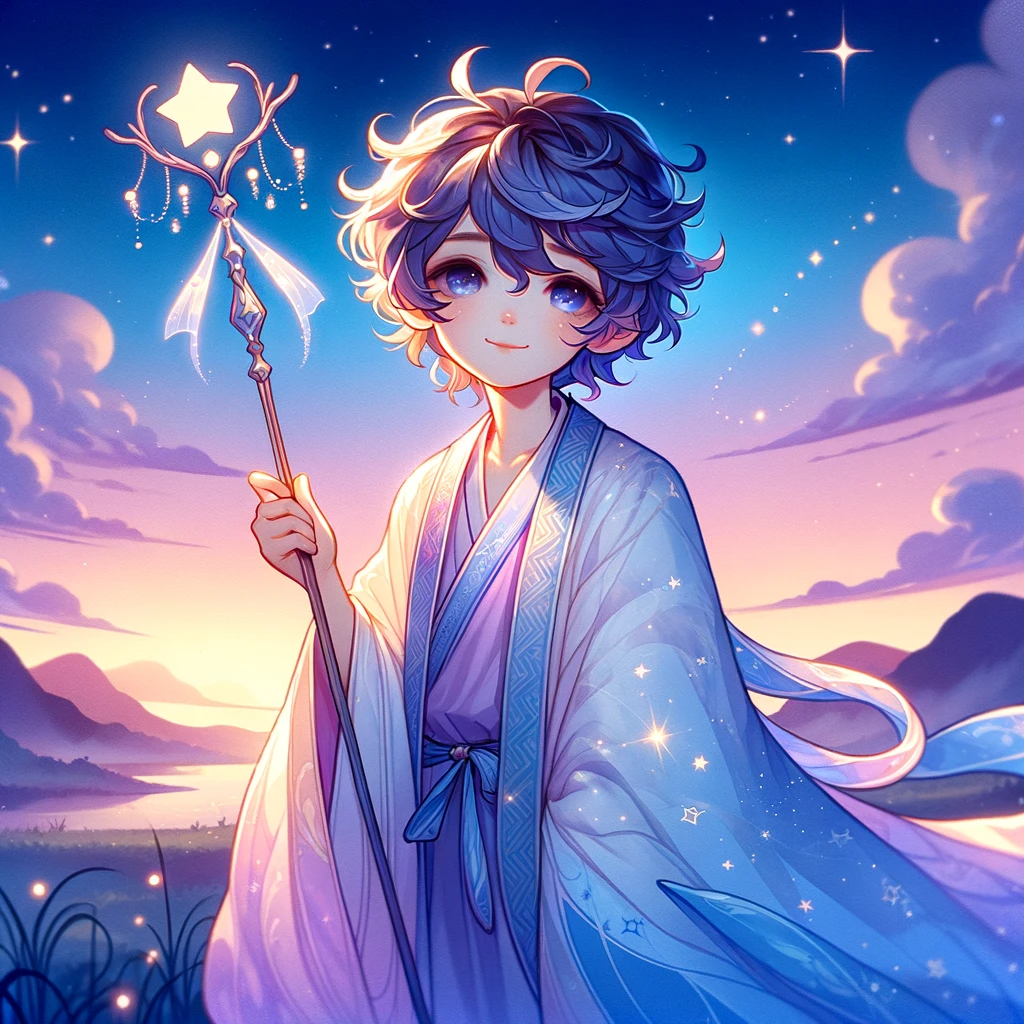}}};
\end{tikzpicture}

\vspace{-14mm}

\begin{abstract}

The high cost of full-parameter fine-tuning (FFT) of Large Language Models (LLMs) has led to a series of parameter-efficient fine-tuning (PEFT) methods. However, it remains unclear which methods provide the best cost-performance trade-off at different model scales. We introduce \textsc{Astraios}, a suite of 28 instruction-tuned OctoCoder models using 7 tuning methods and 4 model sizes up to 16 billion parameters. Through investigations across 5 tasks and 8 different datasets encompassing both code comprehension and code generation tasks, we find that FFT generally leads to the best downstream performance across all scales, and PEFT methods differ significantly in their efficacy based on the model scale. LoRA usually offers the most favorable trade-off between cost and performance. Further investigation into the effects of these methods on both model robustness and code security reveals that larger models tend to demonstrate reduced robustness and less security. At last, we explore the relationships among updated parameters, cross-entropy loss, and task performance. We find that the tuning effectiveness observed in small models generalizes well to larger models, and the validation loss in instruction tuning can be a reliable indicator of overall downstream performance.

\end{abstract}
\begin{figure}[!h]
    \centering
    \includegraphics[width=0.85\textwidth]{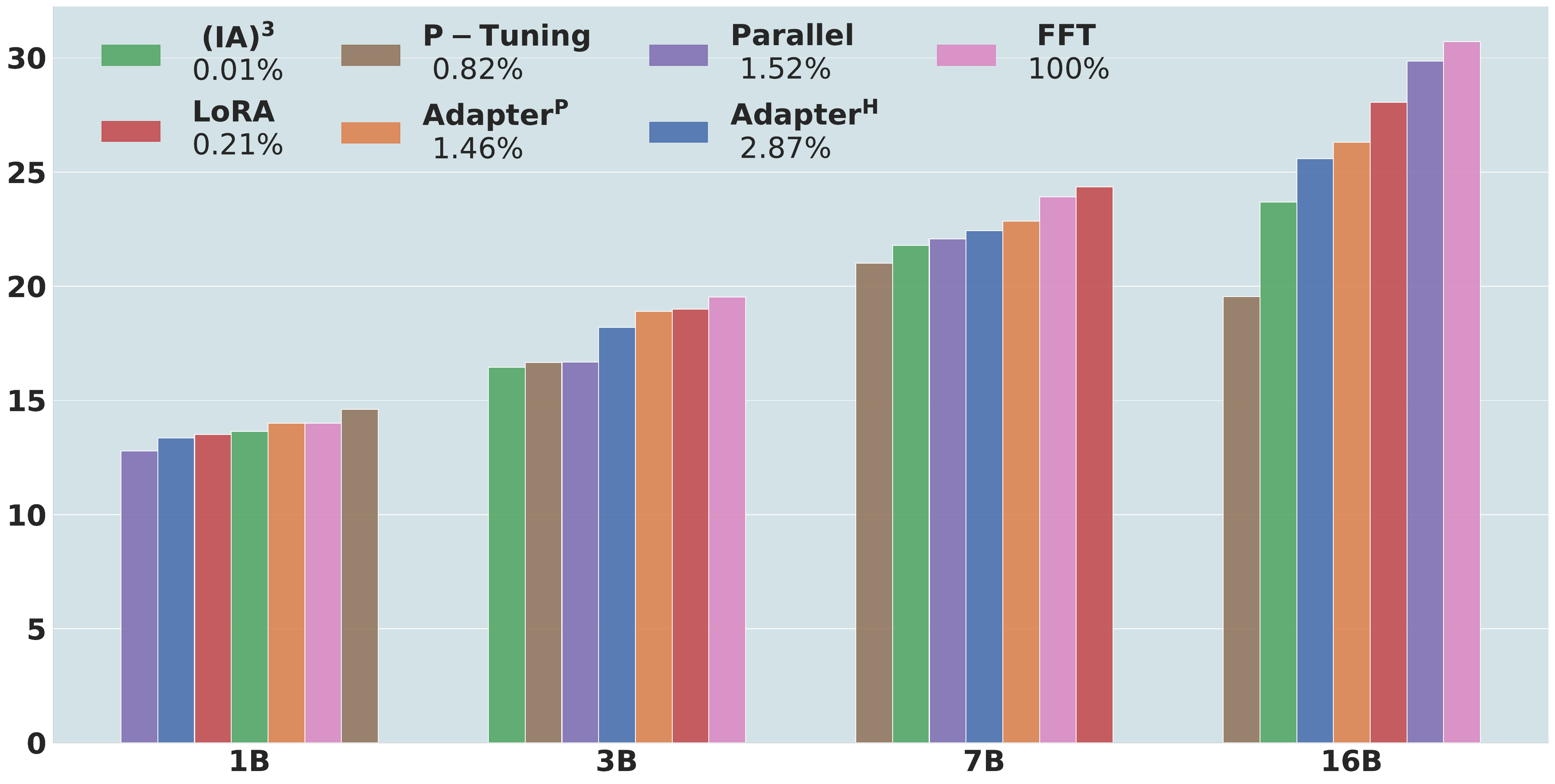}
    \caption{Mean task performance of \textsc{Astraios} models across 5 representative tasks and 8 datasets. We indicate the average percentage of total parameters updated for each PEFT method.}
    \label{fig:main}
\end{figure}

\section{Introduction}

Large language models (LLMs)~\citep{zhao2023survey} trained on Code (Code LLMs) have shown strong performance on various software engineering tasks~\citep{hou2023large}.
There are three main model paradigms: \textit{(A)} Code LLMs for code completion~\citep{nijkamp2022codegen,fried2022incoder,li2023starcoder}; \textit{(B)} Task-specific fine-tuned Code LLMs for a single task~\citep{hou2023large}; and \textit{(C)} Instruction-tuned~\citep{ouyang2022training} Code LLMs that excel at following human instructions and generalizing well on unseen tasks~\citep{wang2023codet5+,muennighoff2022crosslingual}. Recent instruction-tuned Code LLMs, including WizardCoder~\citep{luo2023wizardcoder} and OctoCoder~\citep{muennighoff2023octopack}, have achieved state-of-the-art performance on various tasks without task-specific fine-tuning.
However, with the increasing parameters of Code LLMs, it becomes more expensive to perform full-parameter fine-tuning (FFT) to obtain instruction-tuned models.
In practise, to save computational cost, parameter-efficient fine-tuning (PEFT) have been applied. This training strategy aims to achieve comparable performance to FFT by updating fewer parameters.
While there are many PEFT methods~\citep{ding2022delta}, the predominant PEFT method is still LoRA, which is proposed in 2021~\citep{hu2021lora}. However, there is no empirical evidence showing LoRA remains the best for instruction-tuned code LLMs. In this paper, we investigate instruction-tuned code LLMs with the following research question: \textit{what are the best PEFT methods for Code LLMs?}

Existing analysis on PEFT methods presents several opportunities for further exploration: \textit{(1)} \textbf{Beyond Task-Specific LLMs}. Most prior works~\citep{zhou2022making,ding2023parameter} only focus on the model paradigm \textit{(B)}, where the selected base models are fine-tuned on specific downstream tasks. While these studies provide insights into PEFT methods on task-specific LLMs, the transferability of their findings to the instruction tuning paradigm is unclear. \textit{(2)} \textbf{Diverse Domains}. Studies on PEFT methods tend to evaluate in the predominant domains like vision~\citep{sung2022vl,he2023parameter} and text~\citep{houlsby2019parameter,he2021towards}, leaving other domains like code underexplored. \textit{(3)} \textbf{Inclusive PEFT Methods}. Prior investigations on PEFT mainly consider a limited number of methods, such as adapter-based tuning~\citep{houlsby2019parameter} or reparametric tuning~\citep{aghajanyan2021intrinsic}, which does not capture the full breadth of available methods. \textit{(4)} \textbf{Multidimensional Evaluation}. Previous works only consider limited evaluation on representative downstream tasks~\citep{chen2022revisiting,fu2023effectiveness,ding2023parameter}. We argue that other evaluation dimensions like model robustness~\citep{han2021robust} and output code safety~\citep{weidinger2021ethical, zhuo2023red, pearce2022asleep, dakhel2023github} are also important, especially in the era of LLM agents~\citep{ouyang2022training,openagents2023}. \textit{(5)} \textbf{Scalability}. Most prior PEFT work has only explored LLMs with insufficient scales of model sizes and training time, which makes their scalability questionable~\citep{lester2021power,chen2022revisiting,hu2023llm}.

To explore these identified opportunities further, we introduce \textsc{Astraios}, a suite of 28 instruction-tuned Code LLMs, which are fine-tuned with 7 tuning methods based on the StarCoder~\citep{li2023starcoder} base models (1B, 3B, 7B, 16B). We instruction-tune the models based on the open-source dataset, CommitPackFT from OctoPack~\citep{muennighoff2023octopack}, to balance their downstream capabilities. We utilize PEFT configurations with Hugging Face's best practices~\citep{peft} and integrate a few PEFT methods from recent frameworks~\citep{hu2023llm}. We first inspect the scalability of different tuning methods through the lens of cross-entropy loss during instruction tuning. Specifically, we assess the scales of model size and training time. Our main evaluation focuses on 5 representative code tasks, including clone detection~\citep{svajlenko2021bigclonebench}, defect detection~\citep{zhou2019devign}, code synthesis~\citep{muennighoff2023octopack}, code repair~\cite{muennighoff2023octopack}, and code explain~\citep{muennighoff2023octopack}. We further study the tuning methods from two aspects: \textit{model robustness}~\citep{wang2022recode} and \textit{code security}~\citep{pearce2022asleep}. We assess how well models can generate code based on the perturbed examples and how vulnerable the generated code can be.

The main experimental results can be found in Figure~\ref{fig:main}, where we observe that FFT generally leads to the best downstream performance across all scales. In addition, we find that PEFT methods differ significantly in their efficacy depending on the model scale. At 16B parameters, Parallel Adapter~\citep{he2021towards} and LoRA~\citep{hu2021lora} are the most competitive methods with FFT. Meanwhile, at 1B parameters, they are both slightly outperformed by P-Tuning and (IA)$^3$. Thus, the choice of the PEFT method should be considered along with the model scale at hand.
Nevertheless, LoRA usually offers the most favourable trade-off between cost and performance.

Meanwhile, we also observe that larger PEFT Code LLMs perform better on code generation tasks while they do not show such patterns on code comprehension tasks like clone detection and defect detection. In addition, increasing model size improves generation task performance but exhibits vulnerabilities to adversarial examples and biases towards insecure code. 
Additionally, we investigate the relationships among updated parameters, cross-entropy loss, and task performance. We find that the final loss of small PEFT models can be extrapolated to the larger ones. We also observe strong correlations between final loss and overall downstream task performance. Although the instruction dataset we choose is general and is not directly correlated with the benchmark downstream tasks, we suggest that the performance on such general data can serve as a proxy for the downstream performance.

\section{The \textsc{Astraios} Suite and Benchmark}
In this section, we document our model choices, training configurations, and evaluations in detail for easy reproducing our experimental results in this paper.

\begin{figure}[t]
    \centering
    \includegraphics[width=0.6\textwidth]{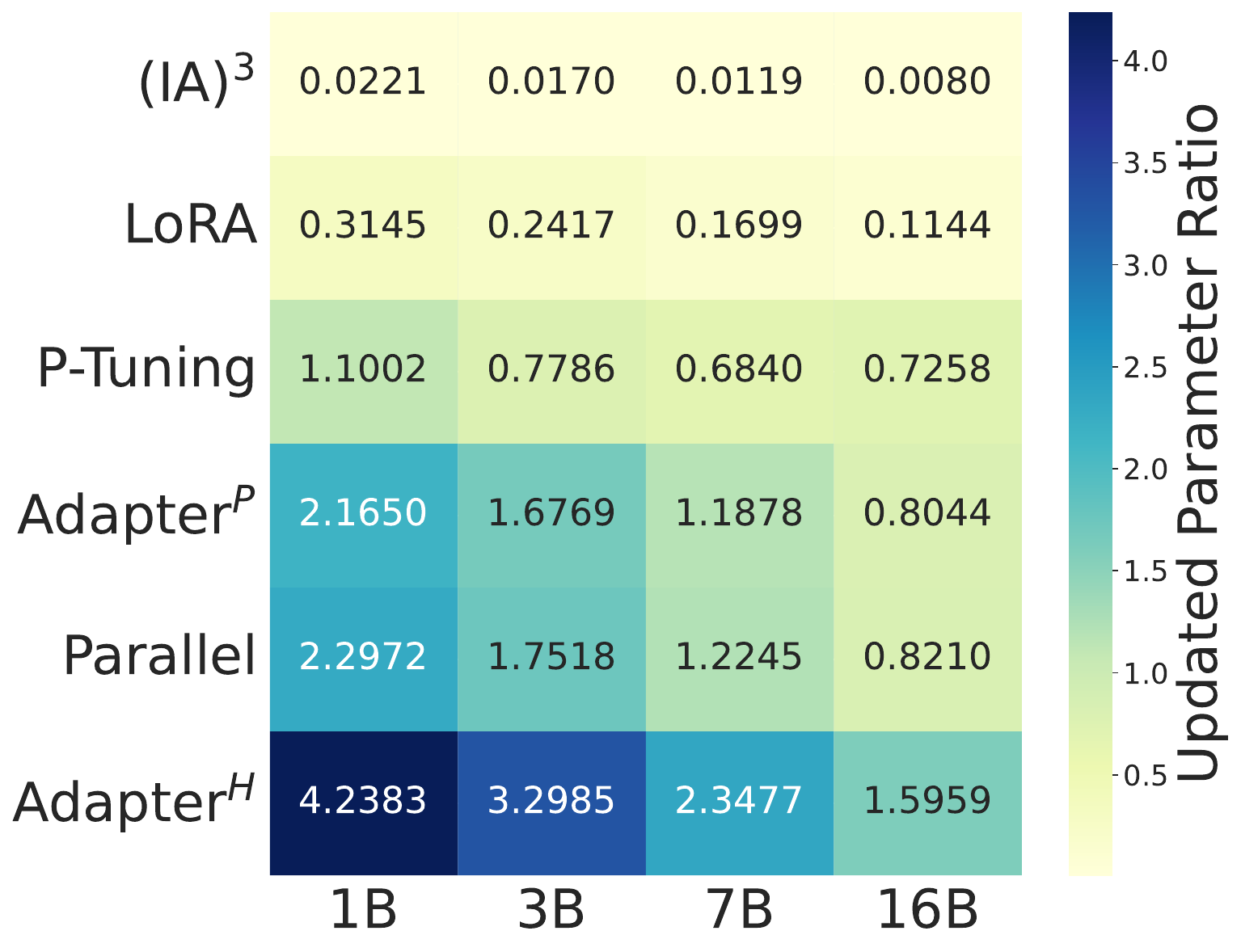}
    \caption{Percentage (\%) of total parameters updated for each PEFT method in \textsc{Astraios} models.}
    \label{fig:param_ratio}
\end{figure}

\begin{table*}[b]
\centering
\caption{Summary of tuning methods and the trainable parameters of different model scales.}
\resizebox{\textwidth}{!}{
\begin{tabular}{llcccc}
\toprule
\textbf{Type} & \textbf{Name}  & \textbf{1B} & \textbf{3B} & \textbf{7B} & \textbf{16B} \\
\midrule
Low-Rank  & LoRA~\citep{hu2021lora} &  3,588,096 & 7,372,800 & 12,472,320 & 17,776,640\\
\midrule
Prompt   & P-Tuning~\citep{liu2023gpt}  & 12,650,496 & 23,882,496 & 50,466,816 & 113,448,960  \\
\midrule
\multirow{4}{*}{Adapter} & (IA)$^3$~\citep{liu2022few}  & 251,904 & 516,096 & 870,912 & 1,239,040   \\
& Adapter$^H$~\citep{houlsby2019parameter}  & 50,331,648&  103,809,024&  176,160,768 & 251,658,240 \\
       & Adapter$^P$~\citep{pfeiffer2020mad}  & 25,165,824 & 51,904,512 & 88,080,384 & 125,829,120 \\
       & Parallel~\citep{he2021towards}  & 26,738,688 & 54,263,808 & 90,832,896 & 128,450,560  \\
\midrule
FFT & FFT  & 1,137,207,296 & 3,043,311,104 & 7,327,263,232 & 15,517,456,384\\ 
\bottomrule
\end{tabular}}
\label{tab:peft}
\end{table*}

\subsection{Model}
\paragraph{Base Model} There are many Code LLMs available that could be a suitable base model. However, some of them are not fully open such as Code-Llama~\citep{roziere2023code}, where the training data is not discussed. To maximize transparency, we select the StarCoder series as our base models. Concretely, four model scales including 1B, 3B, 7B and 16B parameters are selected.

\paragraph{PEFT Model} We focus on three kinds of PEFT methods~\citep{ding2022delta}: \textit{(1)} \textbf{Adapter-based Tuning}~\citep{houlsby2019parameter}: An early approach, which injects small-scale neural modules as adapters to LLMs and only tune these adapters for model adaptation. \textit{(2)} \textbf{Prompt-based Tuning}~\citep{li2021prefix}: It wraps the original input with additional context introducing virtual task-specific tokens without adding layers of modules like adapters. \textit{(3)} \textbf{Intrinsic-rank-based Tuning}~\citep{aghajanyan2021intrinsic}: A representative method is LoRA, which assumes that the change of weights during model tuning has a low rank and thus low-rank changes to the matrices suffice. For all methods, we utilize the implementations in the open-source PEFT library\footnote{\url{https://github.com/huggingface/peft}}~\citep{peft} and the LLM-Adapters work~\citep{hu2023llm} built on top of it. We benchmark 6 PEFT methods, including 4 adapter-based, 1 prompt-based, and 1 intrinsic-rank-based tuning methods as shown in Table~\ref{tab:peft}. We also include FFT for each model size. The ratio of updated parameters of each PEFT method is presented in Figure~\ref{fig:param_ratio}.

\subsection{Instruction Tuning}
\label{sec:training}
\paragraph{Dataset} Following previous work, we select the dataset CommitPackFT+OASST from OctoPack~\citep{muennighoff2023octopack} as the instruction tuning dataset, which helps StarCoder to achieve superior performance.
We note that there could be other choices by utilizing other datasets (e.g., the publicly available dataset CodeAlpaca~\citep{codealpaca}) .
However, they usually focus on a certain aspect of code-related tasks and lack generality to different tasks.

\paragraph{Configuration} We train all models with a sequence length of 2048 tokens, with the batch size as 1, the warm-up step as 12, and the global steps as 200. We set the learning rate as $1 \times 10^{-4}$ for PEFT models and $1 \times 10^{-6}$ FFT models with a cosine scheduler in both cases. For PEFT methods, we use 8-bit-quantized models during training~\citep{dettmers2022gpt3}.

\subsection{Evaluation}

\paragraph{Code Comprehension} To evaluate code comprehension, we select two representative tasks: clone detection and defect detection. Clone detection aims to identify segments of code that are either exact duplicates or structurally similar with variations in identifiers, literals, types, layout, and comments, or even more broadly similar in terms of functionality. Defect detection targets for identifying bugs, vulnerabilities, or antipatterns in code. We select two widely-used datasets from CodeXGLUE benchmark~\cite{lu2021codexglue}: BigCloneBench~\citep{svajlenko2021bigclonebench} and Devign~\citep{zhou2019devign}. As the original BigCloneBench and Devign are designed to evaluate classification models, we prepend additional instructions to prompt the instruction-tuned models to complete such tasks. We follow the evaluation settings of CodeXGLUE and use F1 and Accuracy for BigClone and Devign, respectively. Due to the non-trivial number of test examples in these two datasets, we sample 2,000 from each to save costs. As BigCloneBench and Devign are in the binary classification tasks, we use temperature 0 for model inference to get deterministic outputs.

\paragraph{Code Generation} We use HumanEvalPack~\citep{muennighoff2023octopack}, a benchmark recently proposed that enables easy evaluation of instruction-tuned Code LLMs. The benchmark is structured around three core tasks in code generation, each designed to test different capabilities of the model. The first task, Code Synthesis, involves the model in synthesizing functional code given a function with a docstring detailing the desired code behavior. The second task, Code Repair, challenges the model to identify and fix a subtle bug in an otherwise correct code function, using provided unit tests as a guide. The third and final task, Code Explanation, requires the model to generate a clear and concise explanation for a correctly written code function. For the evaluation on HumanEvalPack, we use its Python and Java splits and compute Pass@1 for each task. We use temperature 0.2 and sample 20 outputs per test example.

\paragraph{Model Robustness} 

Evaluating the robustness of code generation models is crucial in understanding their real-world applicability and reliability. Models that can maintain high-performance levels despite variations and perturbations in input data are more likely to be effective in diverse and dynamic coding environments~\citep{bielik2020adversarial,henkel2022semantic,wang2022recode}. Motivated by such model behaviors, we utilize ReCode~\citep{wang2022recode}, a benchmark framework designed to assess the robustness of Code LLMs. We use HumanEval~\citep{chen2021evaluating} as the base dataset and curated it to mimic natural variations while preserving the semantic integrity of the original inputs. The perturbations cover a range of transformations~\citep{zhuo2023data} on code format, function, variable names, code syntax, and docstrings. These transformations are not arbitrary but represent changes occurring naturally in coding practices. The quality of the perturbed data in ReCode is verified through human evaluation and objective similarity scores, ensuring the relevance and reliability of the dataset for robustness assessment. We use temperature 0.2 and 20 samples per test example for the generation. To compute the level of model robustness, we adopt Robust Pass@k (RP@k) from ReCode and also compute Robust Change@k (RC@k) as follows:

\begin{equation}
  RP@k := \mathbb{E}_x \left[ 1 - \frac{{n - r_c s(x)}}{{\binom{n}{k}}} \right]
\end{equation}
\begin{equation}
RC@k := \left| Pass@k - Robust\ Pass@k \right|
\end{equation}

\paragraph{Code Security} 

One limitation of Code LLMs is their tendency to generate code with potential security vulnerabilities, as various studies have highlighted~\citep{dakhel2023github,asare2023github}. In our work, we aim to empirically examine how PEFT methods can influence the security aspects of Code LLM outputs. We utilize the ``Asleep at the Keyboard'' (AATK) benchmark~\citep{pearce2022asleep}, which includes 89 security-centric scenarios, to provide a comprehensive evaluation across three distinct dimensions: Diversity of Weakness (DoW), encompassing 18 unique vulnerability classes from the MITRE Common Weakness Enumeration (CWE) taxonomy, sourced from the 2021 CWE Top 25 Most Dangerous Software Weaknesses; Diversity of Prompt (DoP), assessing responses to different prompts within the SQL injection vulnerability class; and Diversity of Domain (DoD), involving scenarios in Verilog, a hardware description language. Our analysis predominantly focuses on the DoW axis, comprising 54 scenarios--25 in C and 29 in Python--covering 18 CWEs. This focus is due to the automatic evaluation challenges associated with the other two dimensions. After filtering out scenarios that lack an automated test, we thoroughly examine 40 scenarios, including 23 in C and 17 in Python. We use temperature 0.2 and 20 samples per test example for the generation.

\section{Preliminary Study: Cross-Entropy Loss}
\label{sec:loss}
Cross-entropy loss has been used as the principal performance metric in training LLMs for NLP tasks~\citep{brown2020language,hernandez2021scaling,zhang2022opt}. Most studies on modeling loss focus on either pre-training~\citep{kaplan2020scaling} or FFT~\citep{chung2022scaling}. Previous studies have consistent findings on loss~\citep{kaplan2020scaling, hoffmann2022training, aghajanyan2023scaling}: \textit{The final loss tends to decrease when the training computation (e.g., model sizes, training data and training time) increases}.
These observations indicate that more training time and more trainable model parameters can lead to better alignment with the tuning data. However, there is no systematic investigation for PEFT, especially for Code LLMs. Based on the updated parameters for each tuning method in Table~\ref{tab:peft}, we hypothesize that each PEFT method has a similar trend to previous findings of loss. Inspired by \cite{kaplan2020scaling}, we study the loss change for instruction tuning Code LLMs, varying two factors: \textit{(1)} \textbf{Model Size} (1B - 16B); and \textit{(2)} \textbf{Training Time} (measured in global step, maximum 200 steps). Due to the limited budget, We do not study how the amount of training data may affect the loss.

\begin{figure}[!h]
\centering

\begin{minipage}[b]{0.48\linewidth}
\includegraphics[width=\linewidth]{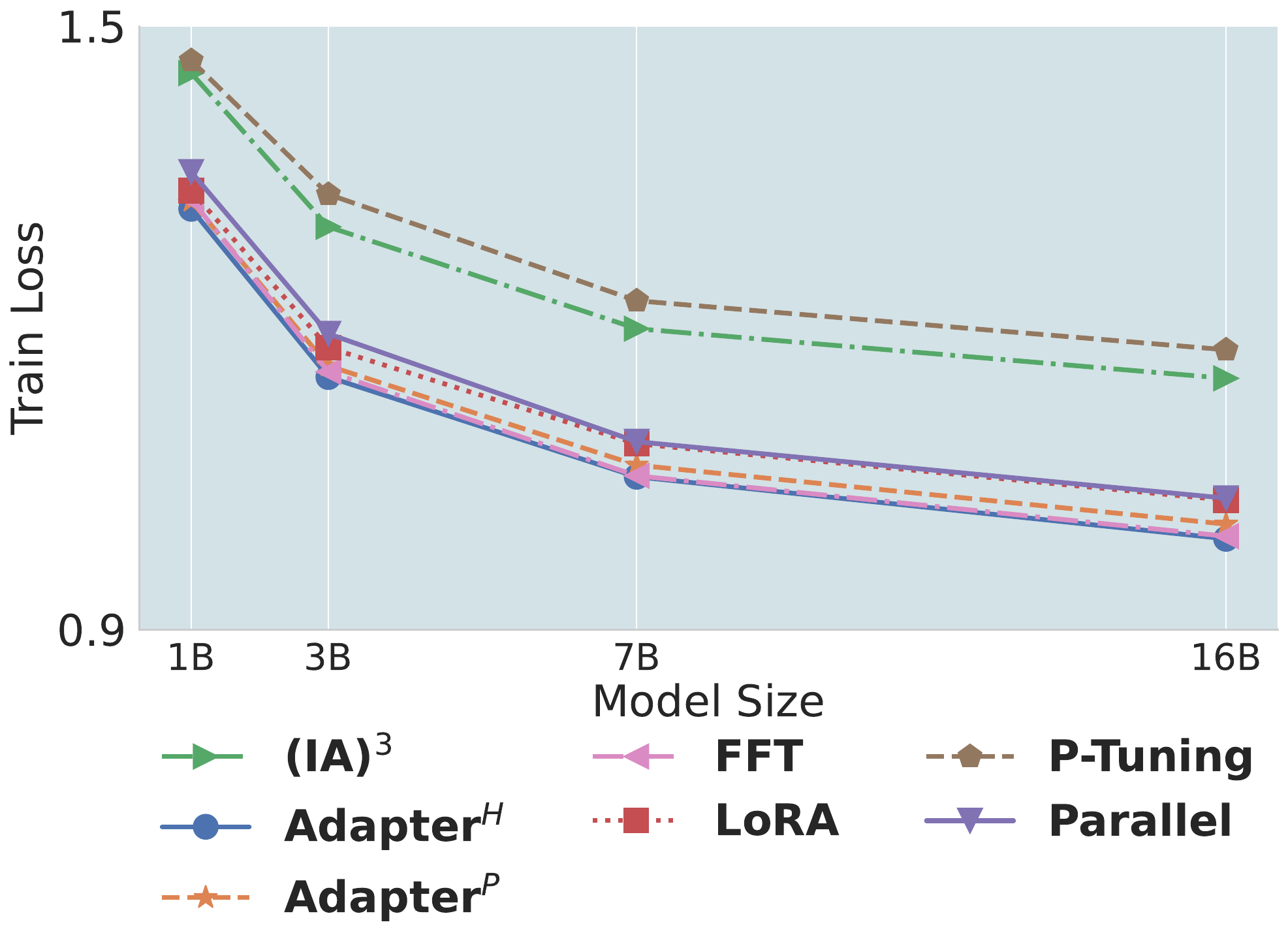}
\end{minipage}
\hfill
\begin{minipage}[b]{0.48\linewidth}
\includegraphics[width=\linewidth]{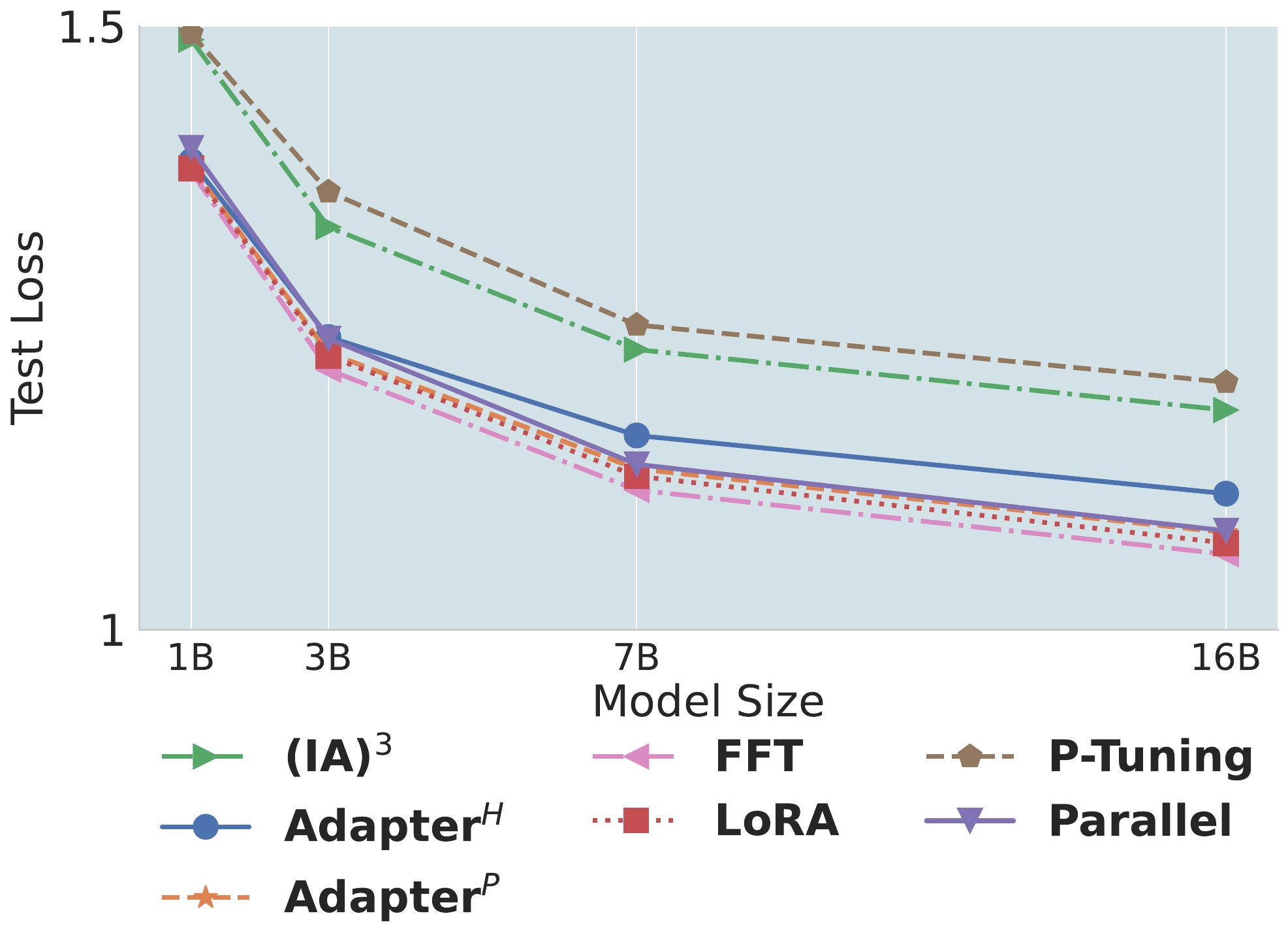}
\end{minipage}
\caption{Final loss across model sizes.}
\label{fig:model_size}
\end{figure}

\paragraph{Model Size Scaling} We present the results of final loss in Figure~\ref{fig:model_size} when varying the model size from 1B to 16B. Our first observation is that train and test loss are well aligned, indicating that the models trained on the selected tuning methods are not overfitted. The second observation is that both train and test loss also strictly decrease when the model size increases. Although these observations are aligned with the aforementioned observations~\citep{kaplan2020scaling, hoffmann2022training}, they show the different scales of loss change, suggesting different tuning methods may require different levels of power. Compared to other tuning methods, FFT demonstrates a slightly better loss performance than PEFT methods like LoRA and Parallel Adapter. As we notice that heavier PEFT methods (which update more parameters) tend to have a better final loss, we hypothesize that more trainable parameters in the model may result in a smaller loss, regardless of how the parameters are updated during training.

\begin{figure}[!h]
\centering

\begin{minipage}[b]{0.48\linewidth}
\includegraphics[width=\linewidth]{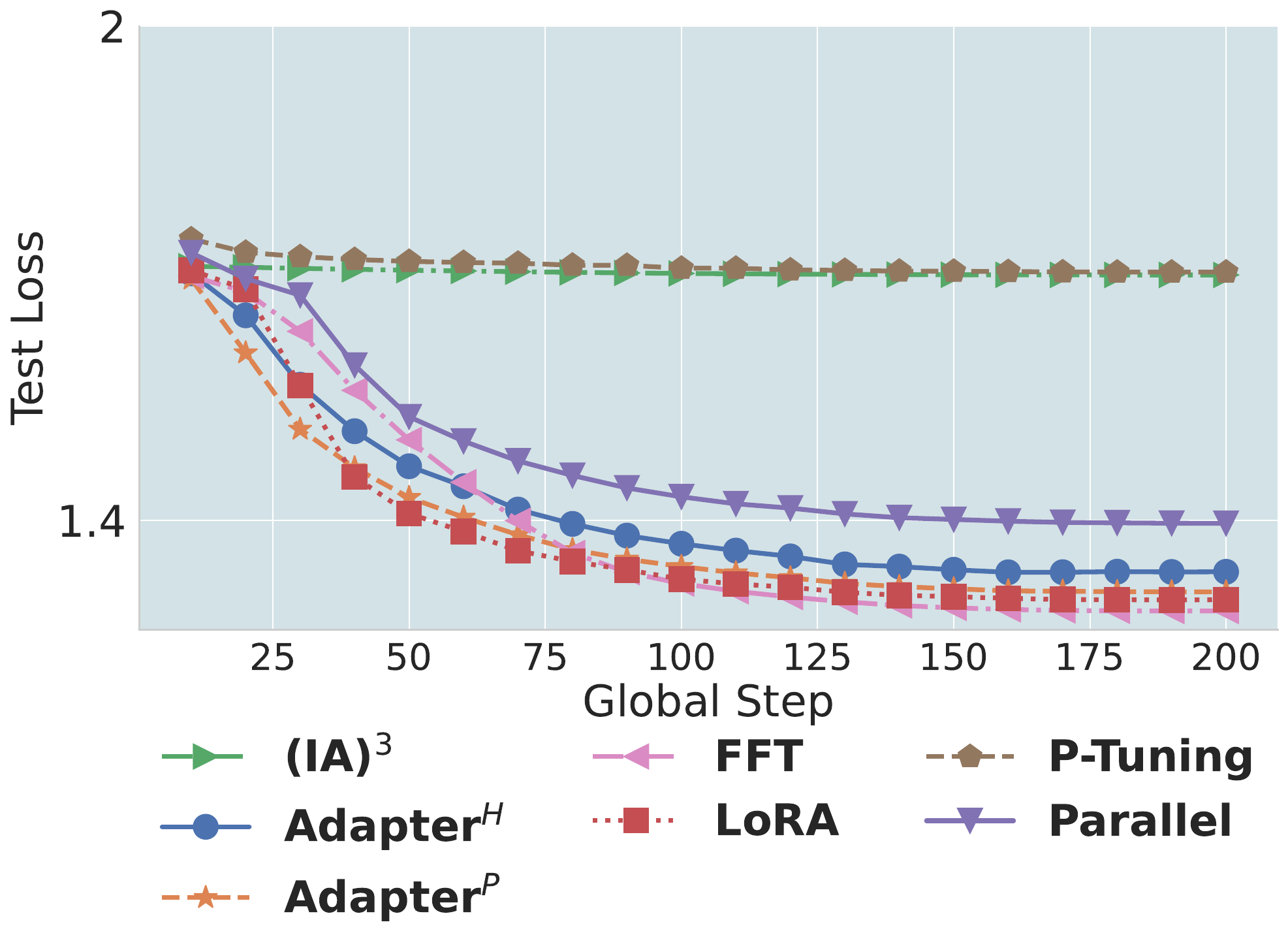}
\caption*{1B \textsc{Astraios} models.}
\end{minipage}
\hfill
\begin{minipage}[b]{0.48\linewidth}
\includegraphics[width=\linewidth]{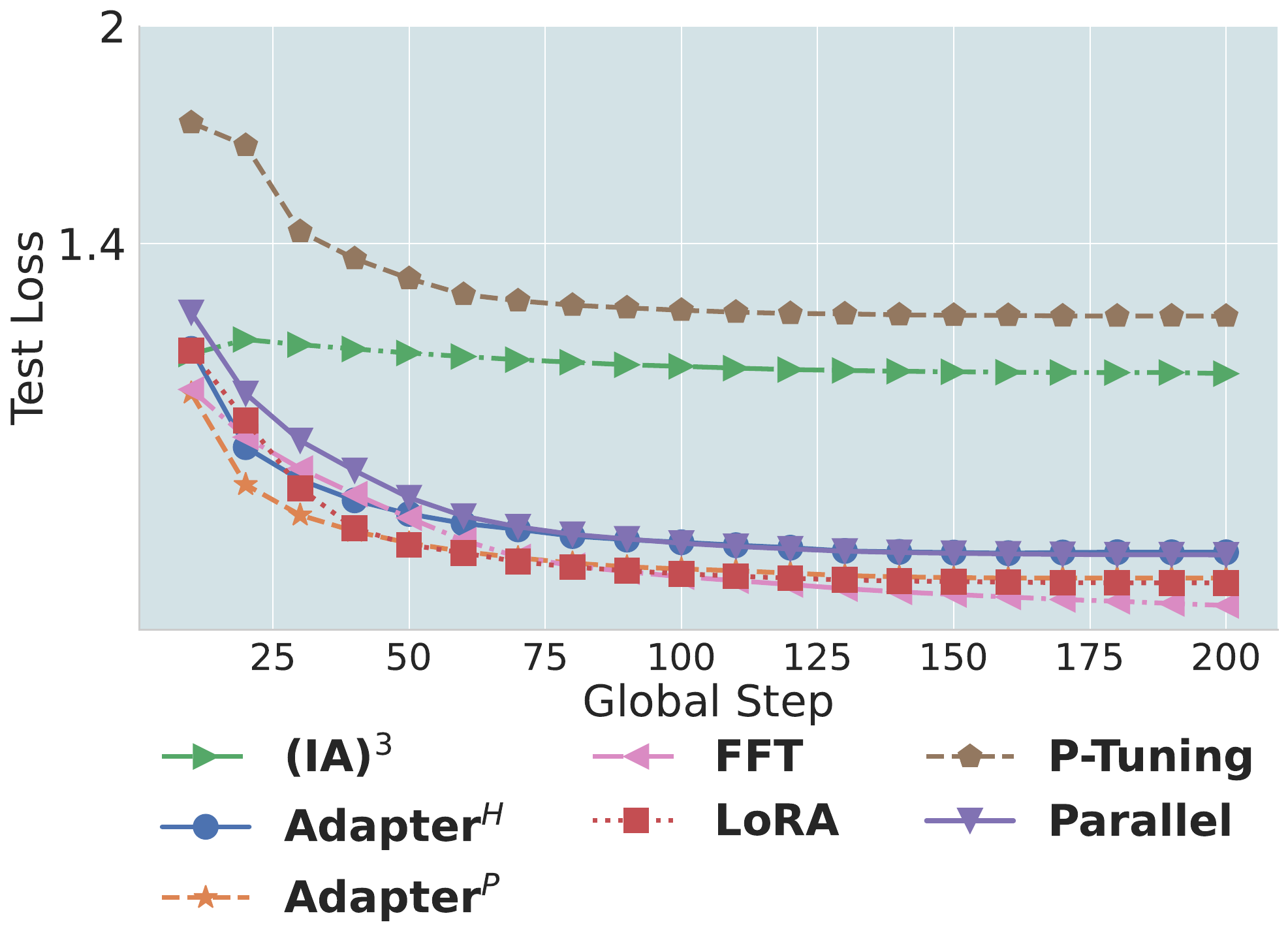}
\caption*{3B \textsc{Astraios} models.}
\end{minipage}

\vspace{1em}

\begin{minipage}[b]{0.48\linewidth}
\includegraphics[width=\linewidth]{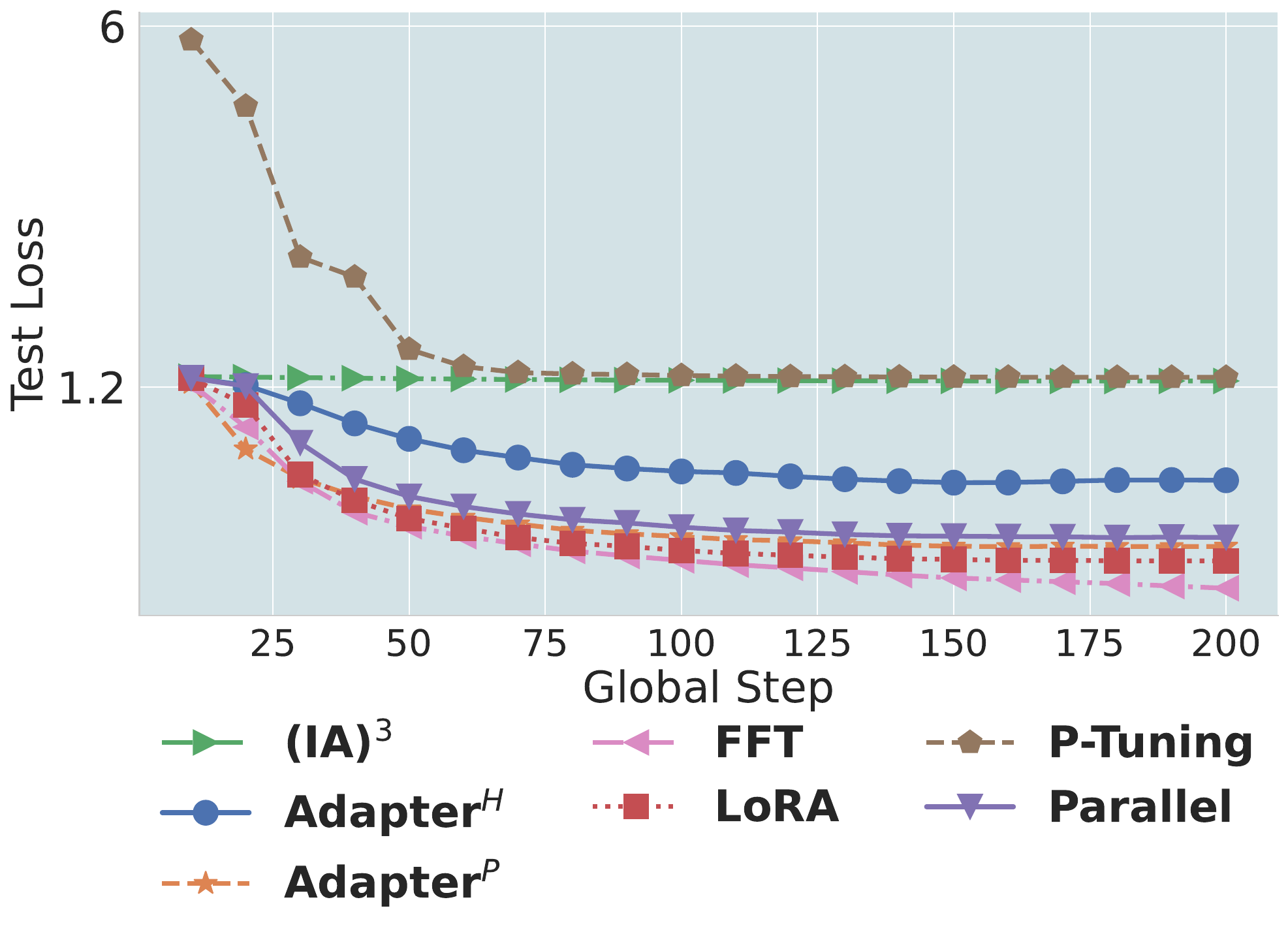}
\caption*{7B \textsc{Astraios} models.}
\end{minipage}
\hfill
\begin{minipage}[b]{0.48\linewidth}
\includegraphics[width=\linewidth]{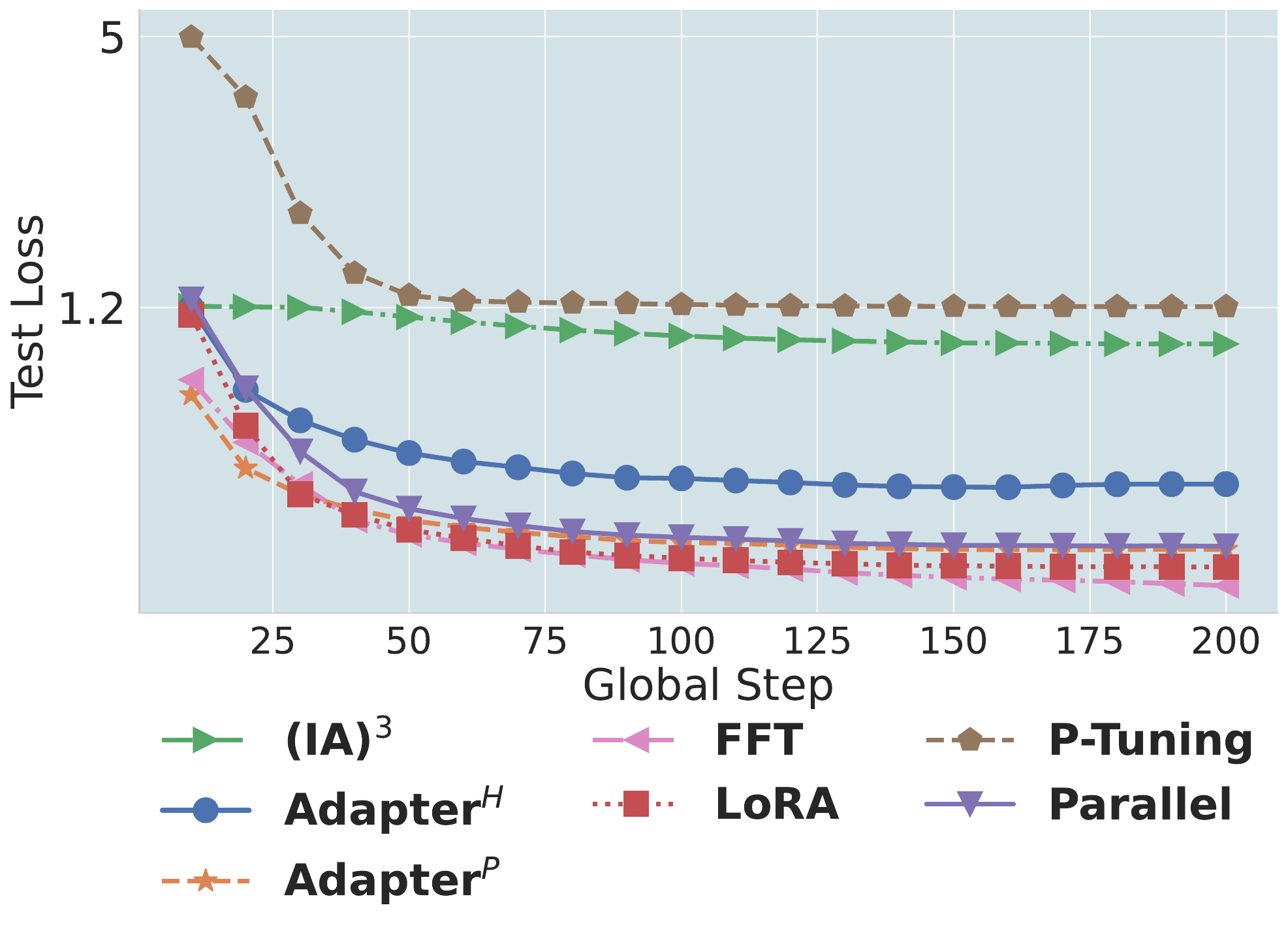}
\caption*{16B \textsc{Astraios} models.}
\end{minipage}

\caption{Test loss of \textsc{Astraios} models across training time measured by \textit{Global Step}.}
\label{fig:training_time}

\end{figure}

\paragraph{Training Time Scaling} We show the changes in test loss on the \textsc{Astraios} when varying the training time in Figure~\ref{fig:training_time}. We notice that the loss continues decreasing when the model is trained longer. Although the loss changes of (IA)$^3$ are consistently insignificant. Notably, the loss of P-Tuning decreases drastically to 50 steps but behaves similarly to other prompt-based methods. In terms of tuning stability, we observe that P-tuning is more unstable than other methods, where the loss change appears to be a non-monotonic pattern. When comparing FFT against PEFT methods, we find that FFT tends to decrease even after 200 steps, while PEFT methods do not show a decreasing trend clearly. We hypothesize that it may be due to the number of updated parameters, where FFT updates the full parameters in the model.

\begin{table}[h]
\centering
\caption{Results of \textsc{Astraios} models on Defect Detection and Clone Detection. The best performance is highlighted in \textbf{bold}. The second best performance is \underline{underlined}.}
\label{tab:code_comprehension}
\resizebox{0.7\textwidth}{!}{
\begin{tabular}{l|cccc|cccc}
\toprule
\multirow{2}{*}{\textbf{Method}} & \multicolumn{4}{c|}{\textbf{Defect Detection}} & \multicolumn{4}{c}{\textbf{Clone Detection}} \\
\cmidrule{2-9}
 & \textbf{1B} & \textbf{3B} & \textbf{7B} & \textbf{16B} & \textbf{1B} & \textbf{3B} & \textbf{7B} & \textbf{16B} \\
\midrule
LoRA             & 44.15            & 44.90             & \underline{49.05} & 31.95 & 9.30             & 12.05             & \underline{14.10}             & 8.80 \\
P-Tuning          & \underline{53.70} & 27.75             & 40.55             & 11.00 & \textbf{19.27}   & \textbf{23.52}    & 13.35             & 3.24 \\
Adapter$^H$         & 45.75            & \underline{45.80} & 46.25             & 41.75 & 8.59             & 8.17              & 12.05             & 8.18 \\
Adapter$^P$         & 45.55            & 46.05    & 46.85             & 27.35 & 8.88             & 8.63              & 12.05             & 9.00 \\
Parallel & 34.50            & 33.50             & \textbf{52.55}    & \underline{42.30} & \underline{9.55} & 8.94              & 10.16             & \textbf{17.21} \\
(IA)$^3$              & \textbf{53.90}   & 33.55             & 37.20             & 23.70 & 8.28             & 11.76             & \textbf{23.19}    & 8.13 \\
\midrule
FFT              & 50.80            & 44.20             & 48.30             & \textbf{43.65} & 8.34             & \underline{12.68} & 8.04              &\underline{12.62} \\
\bottomrule
\end{tabular}}
\end{table}

\section{Main Results: Task Performance}
\label{sec:task}
\begin{figure}[!h]
\centering

\begin{minipage}[b]{0.48\linewidth}
\includegraphics[width=\linewidth]{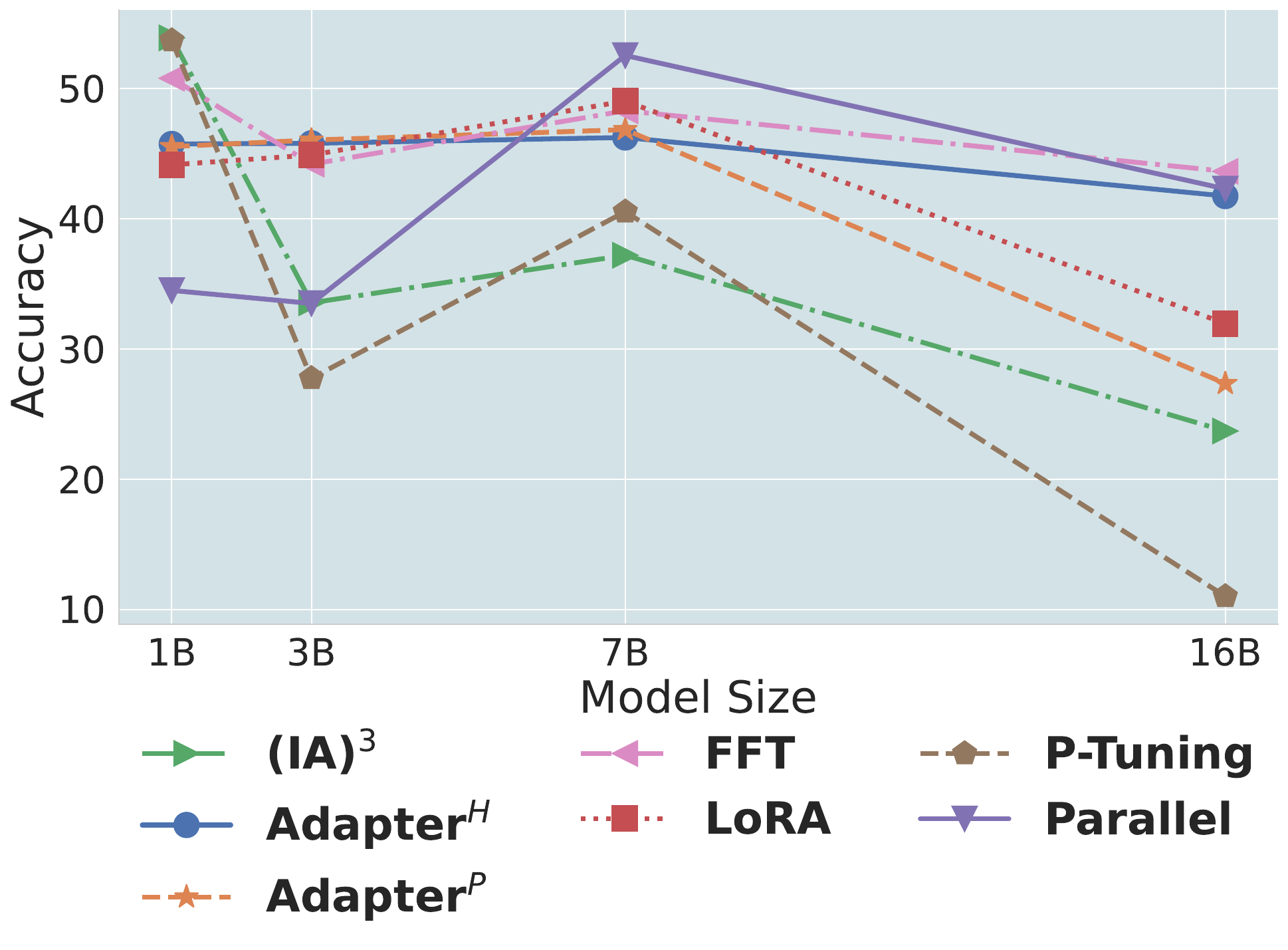}
\caption{Accuracy results of \textsc{Astraios} models on Defect Detection.}
\label{fig:defect_detection}
\end{minipage}
\hfill
\begin{minipage}[b]{0.48\linewidth}
\includegraphics[width=\linewidth]{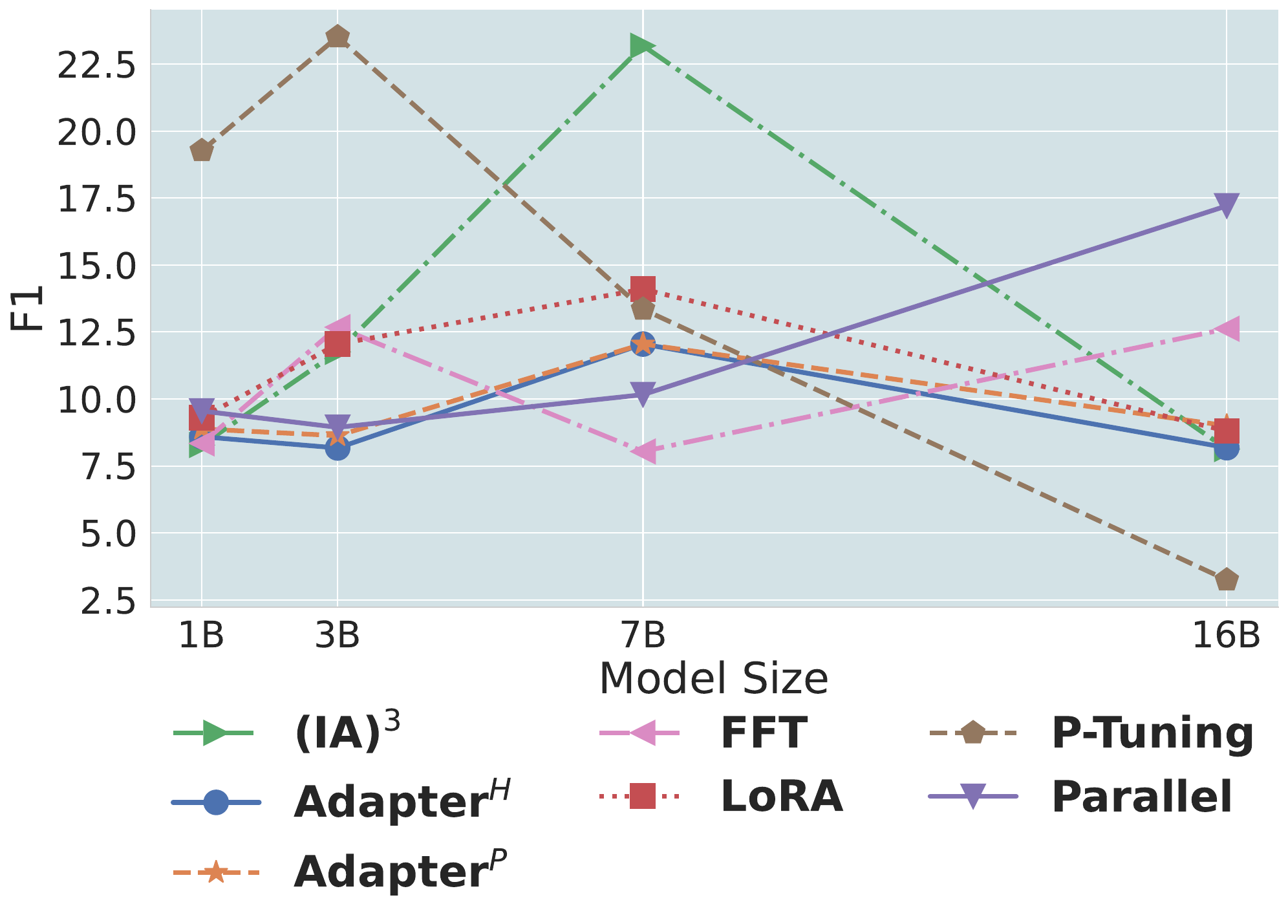}
\caption{F1 results of \textsc{Astraios} models on Clone Detection.}
\label{fig:clone_detection}
\end{minipage}
\end{figure}
As cross-entropy loss only indicates how well Code LLMs can be aligned with the training data, it greatly depends on the specific training content and may not serve as a reliable proxy of performance on various tasks of source code. Therefore, We seek to examine how well selective PEFT methods contribute to task performance in this section. To benchmark the performance, we leverage the representative code downstream tasks: \textit{(1)} Defect Detection, \textit{(2)} Code Clone, \textit{(3)} Code Synthesis, \textit{(4)} Code Repair and \textit{(5)} Code Explanation. For the first two code comprehension tasks, there is no existing study stating that the larger code LLMs result in a better understanding of code. We are the first to study this aspect when varying the model sizes. Regarding the latter three code generation tasks, previous power-law studies~\citep{kaplan2020scaling,hoffmann2022training} have shown that increasing model sizes can also lead to better task performance on generation tasks. We further validate this finding on the PEFT settings.

\paragraph{Code Comprehension} Table~\ref{tab:code_comprehension} shows the results of the two code comprehension tasks when varying the model sizes. Surprisingly, as shown in Figures~\ref{fig:defect_detection} and \ref{fig:clone_detection}, the results of both tasks are not well aligned with the patterns we observe on code generation tasks. All tuning methods consistently behave like the inverse scaling, which has been discussed in \cite{mckenzie2023inverse}. We hypothesize that Code LLMs have not seen enough task-specific training data and cannot generalize to those unseen tasks~\citep{yadlowsky2023pretraining}. As \textsc{Astraios} models are pre-trained on various source code from GitHub repositories for next token prediction and fine-tuned on GitHub commits for code refinement, they may not have a profound understanding of defects and cloned code. 

\begin{table*}[h]
\centering
\caption{Pass@1 results of \textsc{Astraios} models on HumanEvalPack Python and Java splits. The best performance is highlighted in \textbf{bold}. The second best performance is \underline{underlined}.}
\label{tab:humanevalpack}
\resizebox{\textwidth}{!}{
\begin{tabular}{l|l|cccc|cccc|cccc}
\toprule
& \multirow{2}{*}{\textbf{Method}} & \multicolumn{4}{c|}{\textbf{Code Synthesis}} & \multicolumn{4}{c|}{\textbf{Code Repair}} & \multicolumn{4}{c}{\textbf{Code Explanation}} \\
\cmidrule{3-14}
&  & \textbf{1B} & \textbf{3B} & \textbf{7B} & \textbf{16B} & \textbf{1B} & \textbf{3B} & \textbf{7B} & \textbf{16B} & \textbf{1B} & \textbf{3B} & \textbf{7B} & \textbf{16B} \\
\midrule
\multirow{7}{*}{\rotatebox[origin=c]{90}{Python}}
& LoRA             & \textbf{17.26} & \underline{25.37} & \underline{32.01} & \underline{38.08} & 3.29 & 11.16 & \underline{21.74} & \underline{27.50} & \textbf{20.49} & 22.53 & \textbf{25.34} & 30.52 \\
& P-Tuning          & 15.79 & 24.33 & 29.39 & 35.58 & 1.86 & 13.69 & 20.34 & 18.72 & 9.48  & 11.92 & 14.60 & 15.43 \\
& Adapter$^H$         & 15.70 & 23.87 & 28.26 & 33.29 & 3.14 & \textbf{15.55} & \textbf{22.50} & 22.28 & \underline{17.77} & 22.35 & 24.24 & 26.07 \\
& Adapter$^P$         & \underline{17.04} & 24.76 & 30.67 & 34.97 & \underline{3.69} & 12.87 & 19.54 & 26.46 & 16.07 & \textbf{24.05} & 22.87 & 30.67 \\
& Parallel         & 15.98 & \textbf{26.65} & 28.81 & 35.88 & \textbf{4.91} & 8.11  & 16.13 & 26.43 & 19.70 & 23.14 & 23.93 & \textbf{31.10} \\
& (IA)$^3$              & 16.13 & 25.34 & 30.52 & 36.80 & 2.01 & 14.05 & 17.07 & 23.60 & 9.51  & 11.86 & 14.30 & 16.19 \\\cmidrule{2-14}
& FFT              & 16.95 & 25.21 & \textbf{32.38} & \textbf{38.47} & 3.26 & \underline{14.45} & 21.40 & \textbf{29.88} & 15.37 & \underline{23.45} & \underline{26.13} & \underline{30.85} \\
\midrule
\multirow{7}{*}{\rotatebox[origin=c]{90}{Java}}
& LoRA             & 2.84  & 16.52 & \textbf{24.27} & 40.33 & \textbf{3.72} & 5.06  & 13.60 & 30.35 & 7.07  & \textbf{14.33} & 14.70 & \underline{16.86} \\
& P-Tuning          & \textbf{10.67} & 14.73 & 20.73 & 37.19 & 0.00 & \textbf{7.53} & 11.74 & 22.25 & 6.07  & 9.79  & \textbf{17.32} & 13.02 \\
& Adapter$^H$         & 8.99  & 13.45 & 17.53 & 33.41 & 0.12 & \underline{6.89}  & \underline{14.70} & 24.91 & 6.74  & 9.57  & 13.99 & 14.85 \\
& Adapter$^P$         & 10.46 & \underline{16.77} & 21.28 & 33.68 & \underline{3.66} & 6.52  & \textbf{15.40} & \underline{32.07} & 6.65  & 11.62 & 14.15 & 16.28 \\
& Parallel         & 9.60  & 15.91 & 21.59 & 38.56 & 0.49 & 5.09  & 8.87  & 29.39 & \textbf{7.62}  & 12.16 & 14.51 & \textbf{17.93} \\
& (IA)$^3$              & \underline{10.34} & 16.46 & 21.95 & 39.91 & 2.87 & 4.54  & 13.02 & 25.30 & 6.13  & \underline{13.99} & \underline{17.04} & 15.85 \\\cmidrule{2-14}
& FFT              & 10.18 & \textbf{17.04} & \underline{23.87} & \textbf{41.16} & 0.00 & 5.61  & 16.10 & \textbf{32.47} & \underline{7.16}  & 13.60 & 15.12 & 16.62 \\
\bottomrule
\end{tabular}}
\end{table*}

\paragraph{Code Generation} Table~\ref{tab:humanevalpack} demonstrates the performance on three different code generation tasks on the Python and Java splits in HumanEvalPack. Over the six benchmarks, we first observe that FFT results in consistent gains when the model parameters increase.
When examining the PEFT methods, We find they can also provide reasonable performance scalability similar to FFT. Therefore, the lower test loss may lead to better performance across various downstream generation tasks for Code LLMs. However, we notice that the benefit of base model sizes may also differ from tasks and languages. For instance, 1B and 3B models typically underperform in code repair compared to code synthesis. When the model parameters expand to 7B and 16B, their performance across these tasks becomes more comparable.

\paragraph{Overall Performance} To compare the overall task performance of different tuning methods, we compute the mean cumulative scores for each tuning method per model size. We present the rankings in Figure~\ref{fig:main}. We show that FFT remains the best regarding overall task performance, while LoRA and Parallel Adapter are comparable to FFT. However, there is still a huge performance gap between most PEFT methods and FFT, suggesting that they cannot guarantee optimal performance. Regarding the tuning efficiency, we use updated parameters as the metric to summarize two more findings. Firstly, (IA)$^3$ is efficient enough to perform reasonably by updating much fewer parameters than the other PEFT methods. Secondly, we notice that Adapter$^P$ always performs better than Adapter$^H$, even though Adapter$^H$ updates more model parameters. The counter-intuitive observation indicates that Adapter$^H$ may not be worth deploying in real-world practice.

\section{Further Analysis}
In this section, we further study two aspects of Code LLMs beyond task performance. Specifically, we highlight the importance of model robustness and generated code security, which indicate real-world practicality. We tend to understand the trend of model behavior across tuning methods and model sizes.
\subsection{Model Robustness}
\begin{table*}[b]
\centering
\caption{RP@1 and RC@1 results of \textsc{Astraios} models on ReCode. The best performance is highlighted in \textbf{bold}. The second best performance is \underline{underlined}.}
\label{tab:robustness}
\resizebox{\textwidth}{!}{
\begin{tabular}{l|l|cccc|cccc|cccc|cccc}
\toprule
& \multirow{3}{*}{\textbf{Method}} 
& \multicolumn{4}{c|}{\textbf{Format}} 
& \multicolumn{4}{c|}{\textbf{Function}} 
& \multicolumn{4}{c|}{\textbf{Syntax}} 
& \multicolumn{4}{c}{\textbf{Docstring}} \\
\cmidrule{3-18}
& & \textbf{1B} & \textbf{3B} & \textbf{7B} & \textbf{16B}
& \textbf{1B} & \textbf{3B} & \textbf{7B} & \textbf{16B}
& \textbf{1B} & \textbf{3B} & \textbf{7B} & \textbf{16B}
& \textbf{1B} & \textbf{3B} & \textbf{7B} & \textbf{16B} \\
\midrule
\multirow{7}{*}{\rotatebox[origin=c]{90}{Robust Pass}} 
& LoRA & \textbf{28.05} & \textbf{35.98} & 43.29 & \underline{51.22} 
& \textbf{12.80} & 15.24 & \textbf{23.78} & 29.27 
& \textbf{8.54} & \underline{13.41} & \textbf{15.85} & \underline{18.29} 
& \textbf{10.98} & \underline{15.24} & 17.68 & 20.73 \\
& P-Tuning & 18.29 & 29.88 & 39.63 & 48.78 
& 7.32 & \underline{15.85} & 21.34 & 23.78
& 6.71 & 11.59 & 14.02 & 17.68 
& 6.71 & 14.63 & 18.29 & 21.34 \\
& Adapter$^H$ & 10.98 & 34.15 & 40.24 & 46.95 
& 4.88 & 14.02 & 17.07 & 23.78 
& 7.32 & 11.59 & 12.20 & 15.85 
& 6.10 & 12.80 & 14.63 & 17.68 \\
& Adapter$^P$ & 9.76 & \underline{35.37} & \underline{43.90} & 50.00 
& 1.22 & \underline{15.85} & 21.34 & 26.22 
& 4.88 & 12.20 & \underline{14.63} & \underline{18.29} 
& 3.05 & \underline{15.24} & \textbf{19.51} & 20.12 \\
& Parallel & 26.22 & 32.32 & 42.68 & 50.00 
& \underline{10.37} & 11.59 & \underline{21.95} & 26.83 
& \underline{7.93} & 12.80 & \underline{14.63} & 17.07 
& 8.54 & \underline{15.24} & 17.68 & \underline{21.95} \\
& (IA)$^3$ & \underline{26.83} & 33.54 & 42.07 & 50.61 
& \textbf{12.80} & \textbf{17.07} & 21.34 & 26.83 
& \underline{7.93} & 12.20 & \underline{14.63} & 17.07 
& \underline{10.37} & \textbf{15.85} & \underline{18.90} & \textbf{22.56} \\\cmidrule{2-18}
& FFT & 20.12 & \underline{35.37} & \textbf{45.73} & \textbf{53.05} 
& 5.49 & \underline{15.85} & 21.34 & 30.49 
& 7.32 & \textbf{14.63} & \textbf{15.85} & 19.51 
& 6.10 & 14.02 & \underline{18.90} & \textbf{22.56} \\
\midrule
\multirow{7}{*}{\rotatebox[origin=c]{90}{Robust Change}} 
& LoRA            & \underline{10.98} & \underline{14.63} & 15.24 & \underline{15.85} & 4.27 & \underline{6.10} & 4.27 & 6.10 & \textbf{8.54} & \underline{7.93} & 12.20 & 17.07 & \underline{6.10} & \underline{6.10} & \textbf{10.37} & \underline{14.63} \\
& P-Tuning        & 6.10 & 9.76 & 12.80 & \textbf{17.68}  & 4.88 & 4.27 & 5.49 & 7.32 & 5.49 &\textbf{8.54} & \underline{12.80} & 13.41 & 5.49 & 5.49 & 8.54 & 9.76 \\
& Adapter$^H$        & 0.61 & \textbf{15.85} & \underline{15.85} & \underline{15.85} & \underline{5.49} & 4.27 & \textbf{7.32} & 7.32 & 3.05 & 6.71 & 12.20 & 15.24 & 4.27 & 5.49 & \underline{9.76} & 13.41 \\
& Adapter$^P$        & 3.66 & \underline{14.63} & \textbf{17.68} & \underline{15.85} & 4.88 & 4.88 & 4.88 & \underline{7.93} & 1.22 &\textbf{8.54} & 11.59 & 15.85 & 3.05 & 5.49 & 6.71 & 14.02 \\
& Parallel        & \textbf{12.20} & 11.59 & \underline{15.85} & 15.24 & 3.66 & \textbf{9.15} & 4.88 & \underline{7.93} & 6.10 & \underline{7.93} & 12.20 & 17.68 & 5.49 & 5.49 & 9.15 & 12.80 \\
& (IA)$^3$             & \underline{10.98} & 12.80 & 14.02 & 14.63  & 3.05 & 3.66 & \underline{6.71} & \textbf{9.15} & \underline{7.93} &\textbf{8.54} & \textbf{13.41} & \textbf{18.90} & 5.49 & 4.88 & 9.15 & 13.41 \\\cmidrule{2-18}
& FFT             & 7.32 & 14.02 & \textbf{17.68} & 15.24  & \textbf{7.32} & 5.49 & \underline{6.71} & 7.32  & 5.49 & 6.71 & 12.20 & \underline{18.29} & \textbf{6.71} & \textbf{7.32} & 9.15 & \textbf{15.24} \\
\bottomrule
\end{tabular}}
\end{table*}

While the performance on downstream tasks is essential, we argue that the evaluation of model robustness is also necessary to characterize different tuning methods systematically. We therefore consider benchmarking the robustness of code synthesis, one of the most representative downstream tasks of source code.

Table~\ref{tab:robustness} reports each tuning method's worst-case RP@1 and RC@1 of each perturbation category. Among the four types of perturbation, all models perform the worst on syntax transformation, confirming the findings in ~\cite{wang2022recode}. Furthermore, RP@1 per tuning method increases when the model size is scaled up, indicating the generation capability is consistently improved. We noticed that FFT may not perform better than other PEFT methods on smaller models, such as 1B and 3B. However, it results in the best RP@1 on larger models like 16B. By comparing different model sizes, we observe that RC@1 consistently increases when the model gets bigger, indicating that larger models will be less robust.
\begin{figure}[H]
    \centering
    \includegraphics[width=0.9\textwidth]{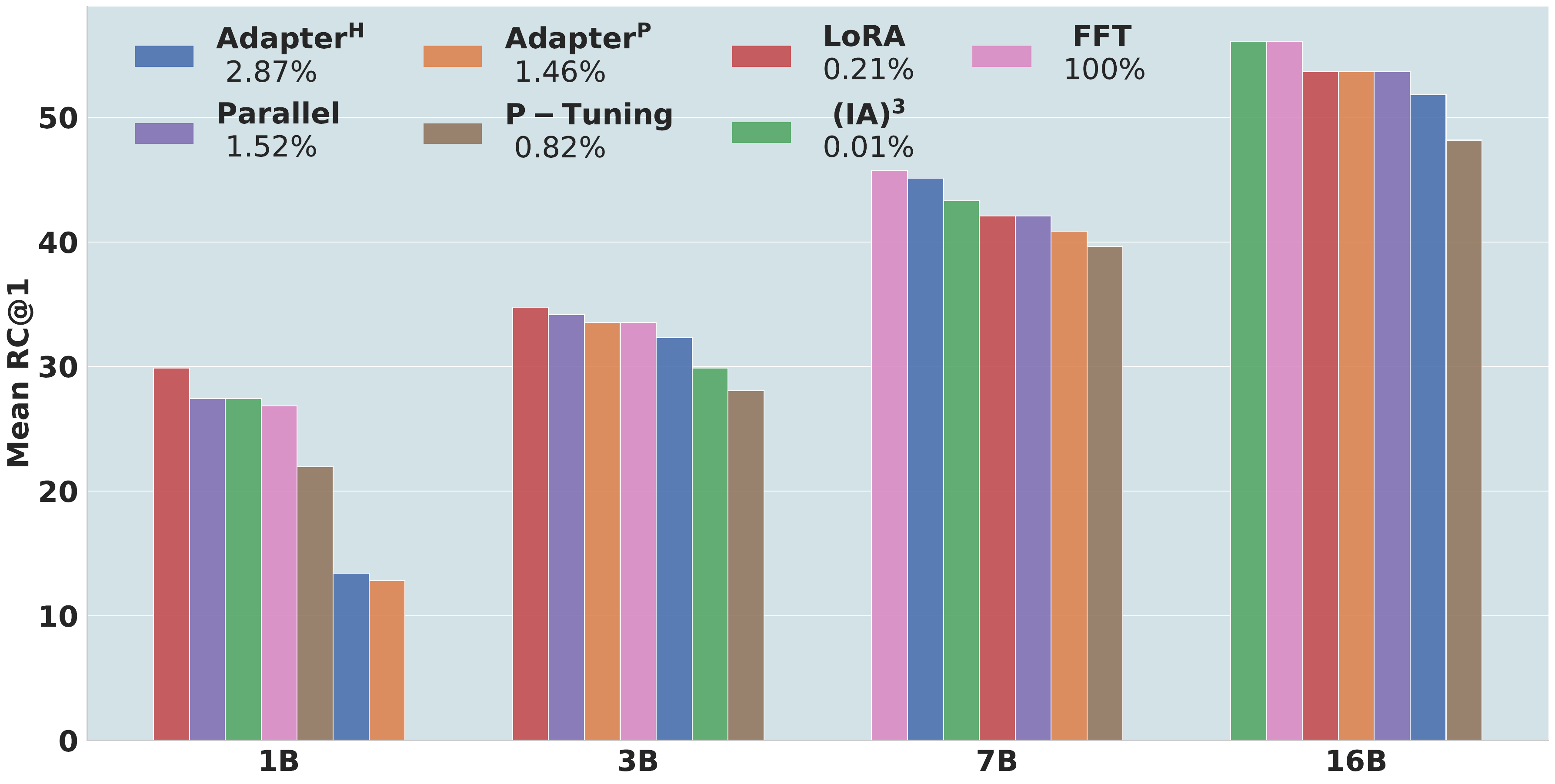}
    \caption{Mean RC@1 of \textsc{Astraios} on ReCode. Lower RC@1 indicates better robustness. We indicate the percentage of total parameters updated for each PEFT method.}
    \label{fig:recode}
\end{figure}
To rank among the tuning methods through the lens of robustness, we compute the mean RC@1 similar to Section~\ref{sec:task} and illustrate in Figure~\ref{fig:recode}. We observe that FFT and LoRA do not show strong robustness. Instead, adapter-based tuning seems more robust while having comparable performance to FFT, which is similar to what \citet{han2021robust} have found in NLP tasks.

\subsection{Code Security}

\begin{table}[b]
\caption{Valid and Insecure rates of \textsc{Astraios} models on AATK benchmark. We note that the insecure rate is calculated based on the valid programs. The best performance is highlighted in \textbf{bold}. The second best performance is \underline{underlined}.}
\label{tab:aatk}
\centering
\resizebox{0.65\textwidth}{!}{
\begin{tabular}{l|cccc|cccc}
\toprule
\multirow{2}{*}{\textbf{Method}} & \multicolumn{4}{c|}{\textbf{Valid\% ($\uparrow$)}} & \multicolumn{4}{c}{\textbf{Insecure\% ($\downarrow$)}} \\
\cmidrule{2-9}
 & \textbf{1B} & \textbf{3B} & \textbf{7B} & \textbf{16B} & \textbf{1B} & \textbf{3B} & \textbf{7B} & \textbf{16B} \\
\midrule
LoRA             & \underline{85.9} & 89.1 & 75.9  & \textbf{87.1}  & \underline{23.1}& 26.2& \textbf{20.9} &35.0\\
P-Tuning       & 70.1 & 68.6 & \underline{86.8} &82.0   & 32.8 & 25.9 & 28.1 &34.5\\
Adapter$^H$        & 84.5 & 90.9 & \textbf{87.5} & \underline{86.8}  & 29.0 & 26.0 & 31.9 &34.1\\
Adapter$^P$      &  83.9&  92.1&  82.8&  86.3    & 31.7 & \underline{25.2} & 26.6 &37.8\\
Parallel &  \textbf{88.9}&  \textbf{94.1}&  70.0 & 86.0 & 30.2 & \textbf{19.3} & 22.3 &\underline{32.6}\\
(IA)$^3$    & 78.0& 62.1& 77.4 &86.6  & 34.8 & \underline{25.2} & 23.1 & \textbf{30.4}\\
\midrule
FFT   &82.9& \underline{93.6}& 80.1 &84.1   & \textbf{22.6} & 27.4 & \underline{21.2} &38.3\\
\bottomrule
\end{tabular}}
\end{table}

Previous studies~\citep{dakhel2023github,asare2023github}. have shown that Code LLMs can generate code with security vulnerabilities, which can be exploited by malicious users. However, few studies have studied different tuning methods from the output security perspective. In this experiment, we intend to understand how tuning methods affect the capability to generate secure code on AATK benchmark.

We follow the original setting in \cite{pearce2022asleep} and compute the valid and insecure rates, which are illustrated in Table~\ref{tab:aatk}. When comparing the valid rate of PEFT methods, it does not show better performance when the model size increases, indicating that current models may not learn the program validity intrinsically. However, we observe that the changes in the insecure rate show that larger models are more likely to generate insecure code. This observation suggests that the growth of learning capability can result in learning more data, including insecure programs. The study on the insecure rate among tuning methods further shows that FFT and LoRA are still better than the other tuning methods regarding the security level. While the other methods have a similar insecure rate, P-Tuning may have more chances to generate less secure programs, which may not be suitable for deploying in security-sensitive scenarios.
\section{Discussion}
In this section, we seek to conduct a preliminary analysis of the performance of Code LLMs through the lens of updated parameters. Specifically, we ask two questions: \textit{(1)} \textit{What is the relationship between the updated parameters and cross-entropy loss?}; and \textit{(2)} \textit{Can we utilize the performance of loss to predict the task performance of Code LLMs?}. 

\begin{figure}[!h]
\centering
\begin{minipage}[b]{0.48\linewidth}
\includegraphics[width=\linewidth]{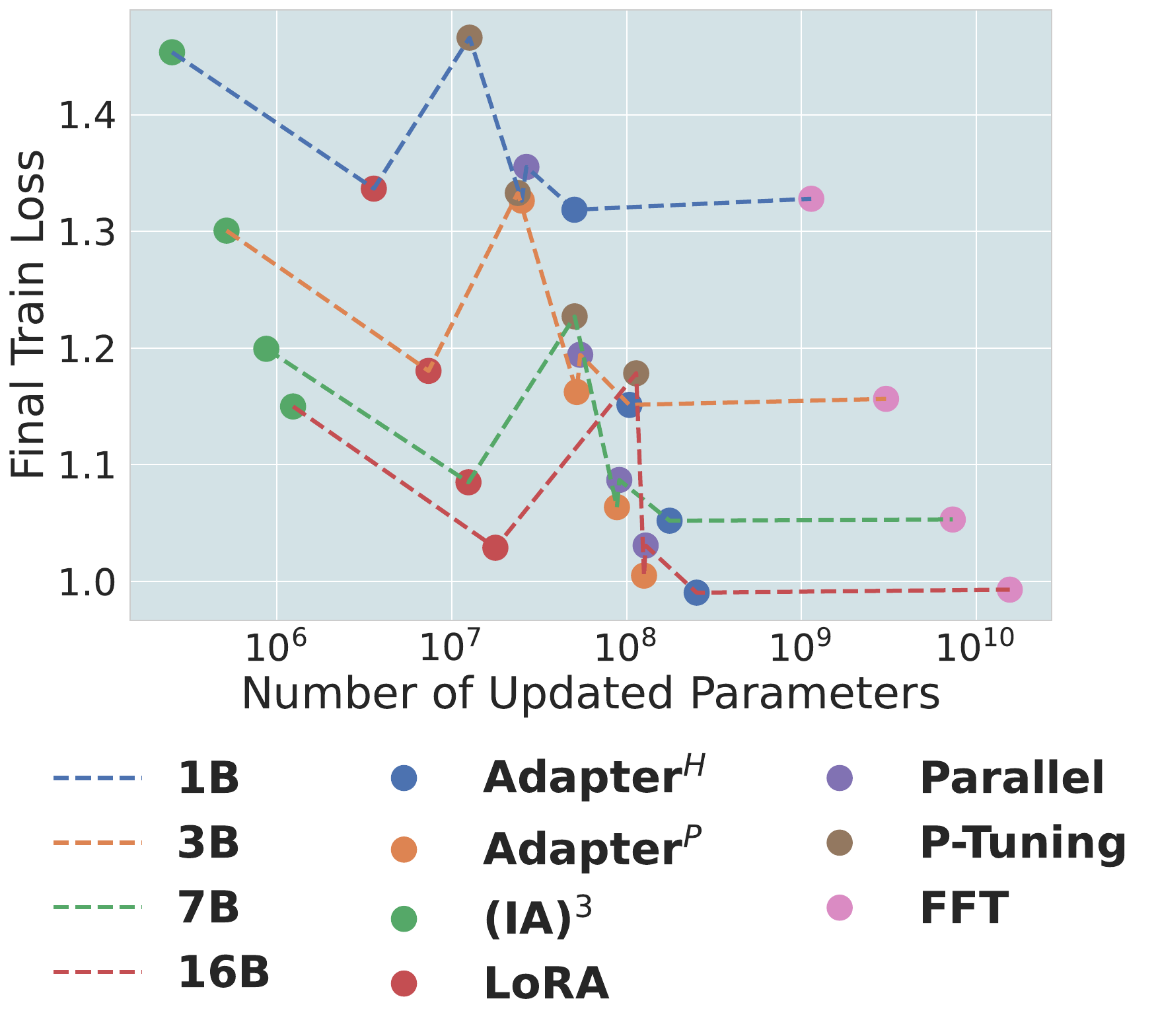}
\end{minipage}
\hfill
\begin{minipage}[b]{0.48\linewidth}
\includegraphics[width=\linewidth]{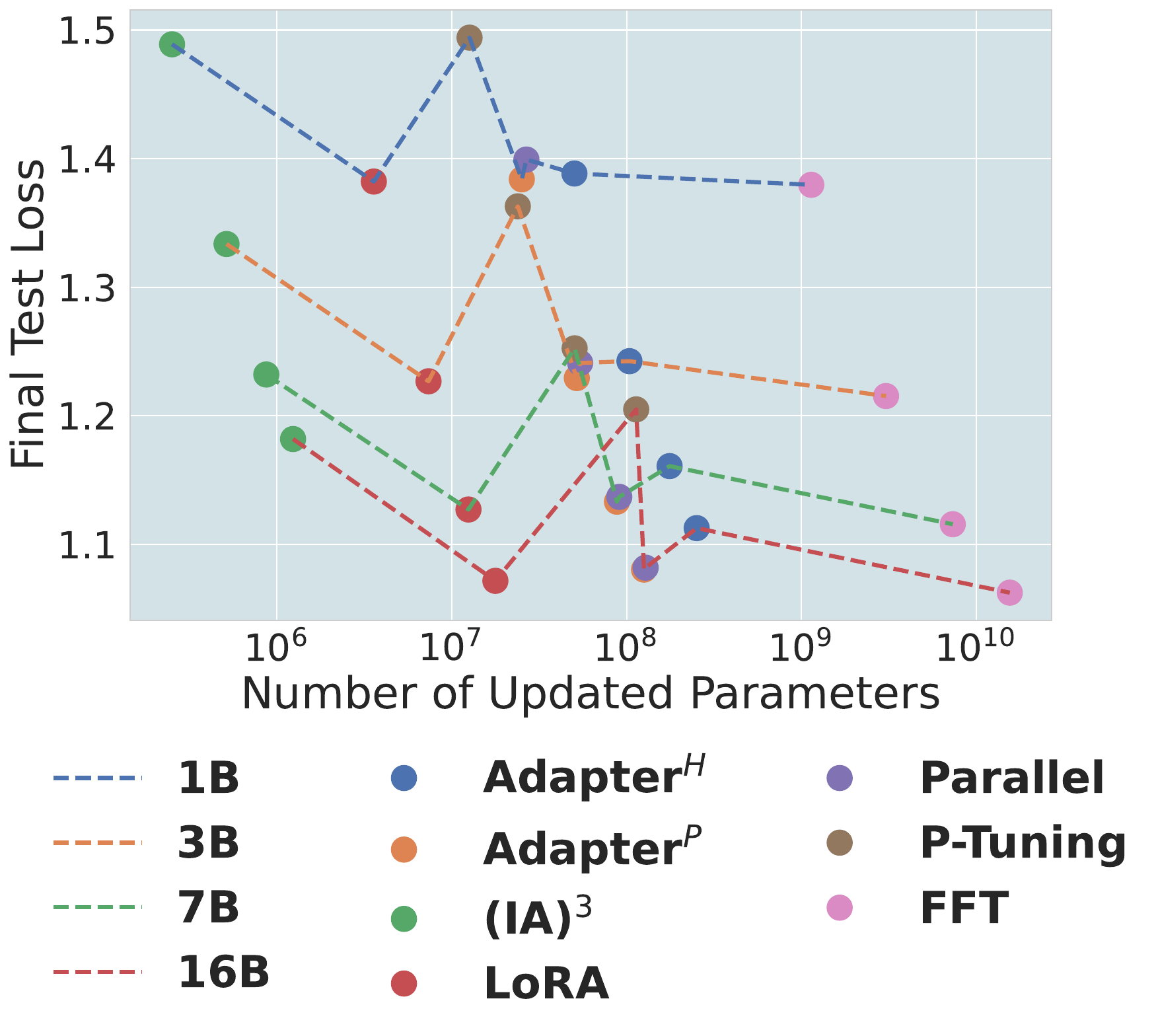}
\end{minipage}
\caption{Relationships between cross-entropy loss and the number of updated parameters.}
\label{fig:para_loss}
\end{figure}

\paragraph{Loss of small models can be projected to larger ones.} The relationship between the updated parameters of \textsc{Astraios} models and their final loss is analyzed in Figure~\ref{fig:para_loss}. Our analysis does not reveal a consistent pattern across different model sizes when it comes to the correlation between model loss and updated parameters. However, an interesting finding is the consistency in relative loss performance across different model sizes when comparing various tuning methods. This consistency suggests that the improvements achieved by each tuning method are likely to be similar regardless of the model's size. Therefore, the loss observed in smaller models, when tuned with different methods, can be a useful predictor for the performance of the larger models.

\begin{figure}[!h]
\centering
\begin{minipage}[b]{0.48\linewidth}
\includegraphics[width=\linewidth]{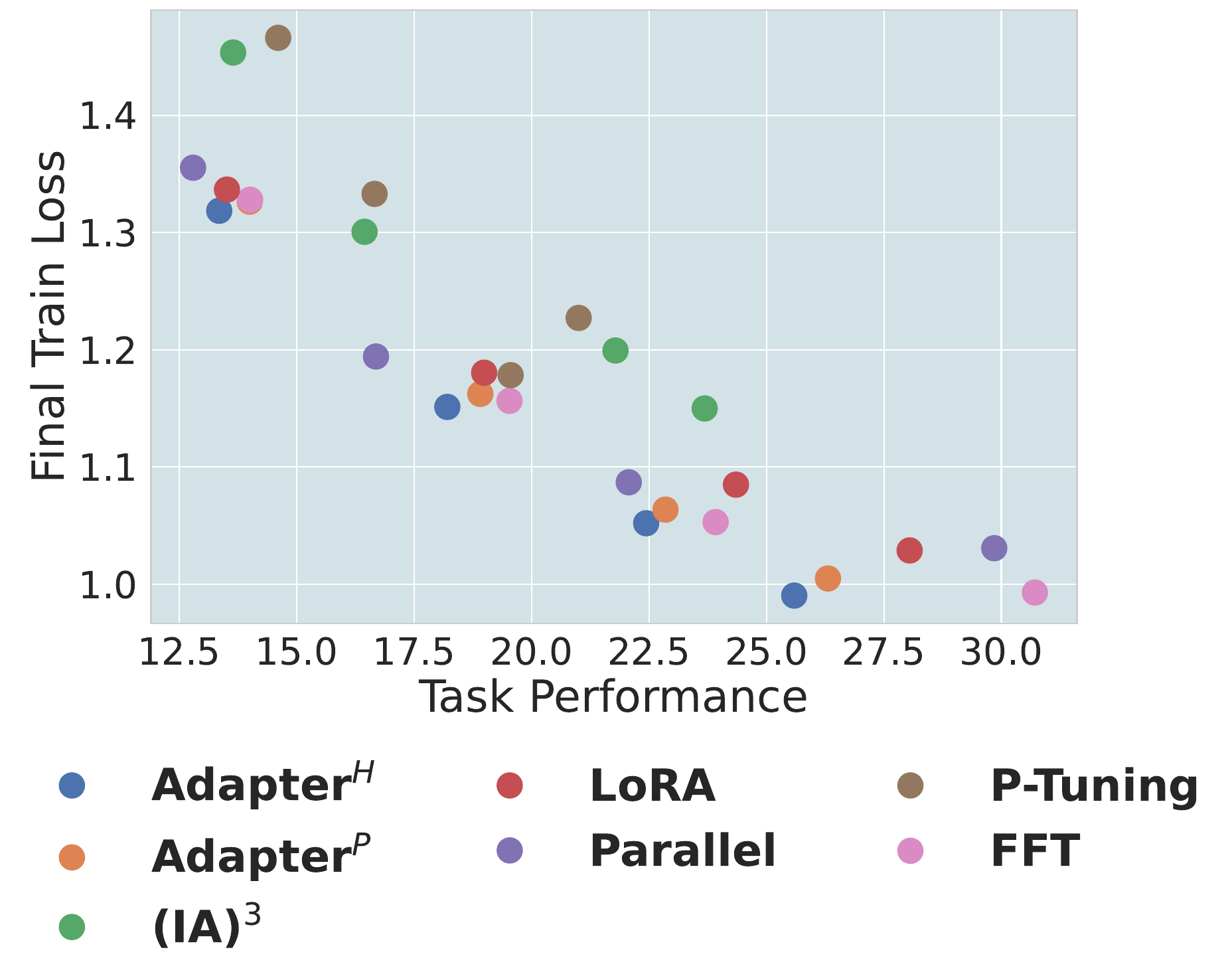}
\end{minipage}
\hfill
\begin{minipage}[b]{0.48\linewidth}
\includegraphics[width=\linewidth]{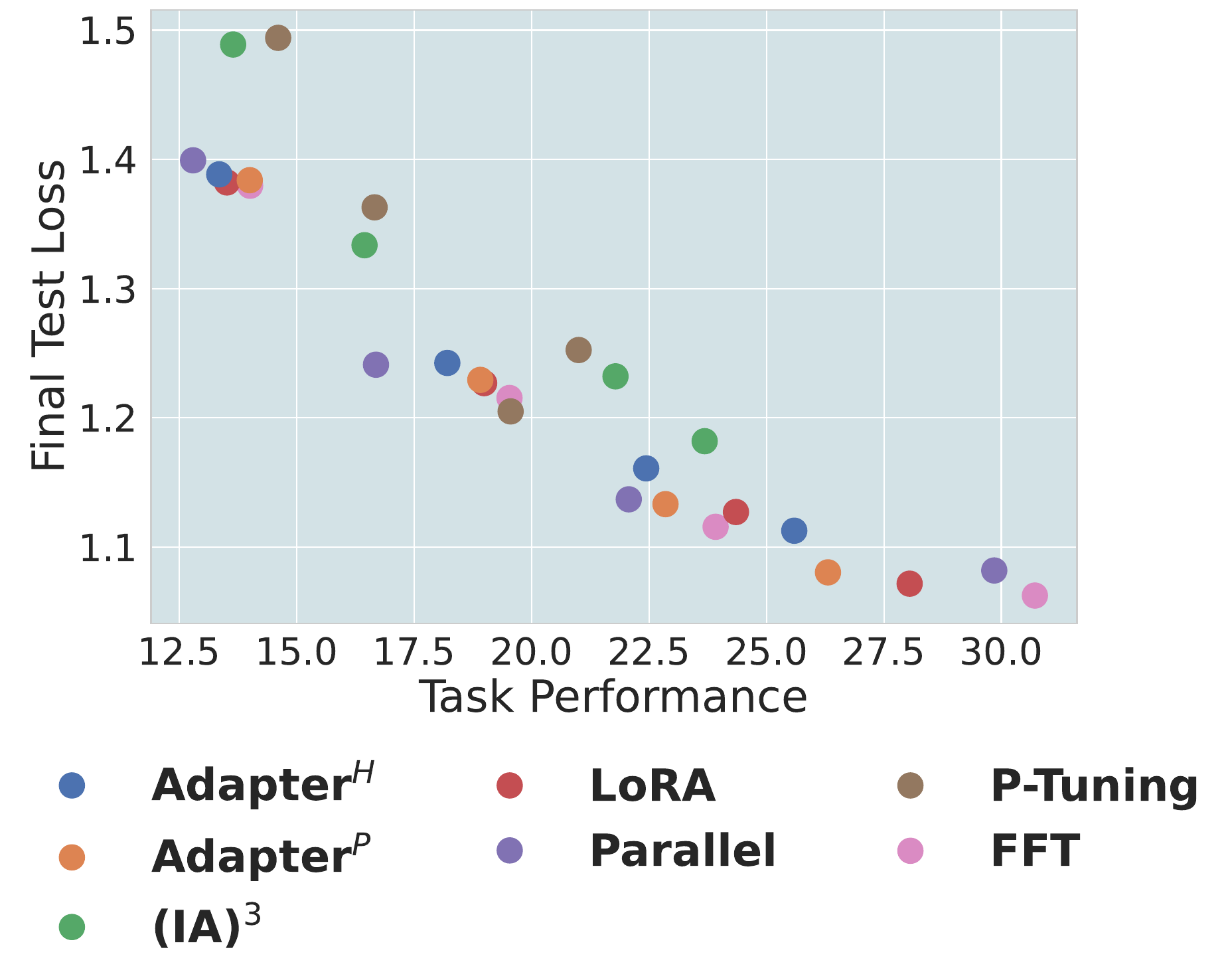}
\end{minipage}
\caption{Relationships between cross-entropy loss and overall task performance.}
\label{fig:loss_downstream}
\end{figure}

\paragraph{Instruct-tuning loss is a strong predictor of downstream performance.} Assuming that the model has been instruction-tuned already but not yet done for the evaluation, we seek to understand if we can utilize such loss to predict its performance on downstream tasks.
Despite our instruction data being derived from general sources like GitHub commits and broad NLP domains, which are not directly aligned with the downstream tasks discussed in Section~\ref{sec:task}, we find some strong correlations.
Motivated by the aforementioned scenario, we aggregate all the data points of mean task performance and their corresponding final loss in Figure~\ref{fig:loss_downstream}. We observe that the models with lower loss generally have better overall performance on downstream tasks. Specifically, the pattern is stronger on test loss than on train loss. We explain by the fact that the models do not learn to fit the test split and can present a more accurate determination of their actual performance. Our observation suggests that general instruction data can work as a good proxy of downstream tasks in Code LLMs, similar to the prior findings in NLP \citep{anil2023palm,wei2023skywork}.

\section{Related Work}

\paragraph{Code Large Language Models} Many base Code LLMs have been proposed recently~\citep{chen2021evaluating,nijkamp2022codegen,fried2022incoder,allal2023santacoder,zheng2023codegeex,li2023starcoder,roziere2023code} mostly targeting code completion. With the help of these base Code LLMs, there have been extensive studies fine-tuning task-specific Code LLMs to perform software engineering tasks like automatic program repair~\citep{xia2023conversational,xia2023automated}, code translation~\citep{pan2023understanding} and code summarization~\citep{wang2023codet5+,wang2022no}. Later, a series of works has been proposed for instruction-tuning the base Code LLMs~\citep{luo2023wizardcoder,shen2023pangu,muennighoff2023octopack,bai2023qwen}, aiming to enhance the generalization capabilities of these models on diverse tasks. As fine-tuning Code LLMs with full parameters is costly, most models have been tuned with LoRA~\citep{hu2021lora}, a parameter-efficient tuning method. In this work, we seek to answer how good LoRA is and if there are other comparable tuning methods.

\paragraph{Model Analysis Across Scales} Understanding why and how neural models behave is crucial for developing more advanced ones. Existing studies have investigated predictable patterns in the behavior of trained language models across scales~\citep{kaplan2020scaling,henighan2020scaling,hernandez2021scaling,hoffmann2022training,wei2022inverse,muennighoff2023scaling,xia2022training} and their learning dynamics~\citep{mcgrath2022acquisition,tirumala2022memorization,biderman2023pythia}. However, they either focus on pre-training or task-specific full-parameter fine-tuning. There is no attempt to understand the mechanism of parameter-efficient instruction tuning. In this paper, we work on this perspective and analyze Code LLMs~\citep{wan2022they,troshin2022probing,zhuo2023pop}.

\section{Conclusion}
This work studies the parameter-efficient instruction-tuning of Code LLMs. We introduce a model suite consisting of 28 instruction-tuned OctoCoder across scales and PEFT methods. We characterize the tuning methods on representative downstream tasks, model robustness, and output security, highlighting the importance of understanding these models via comprehensive evaluation. We also discuss the relationships among updated parameters, cross-entropy loss, and task performance. We hope these analyses will inspire further follow-up work on understanding the mechanism of tuning methods and developing new approaches.

\section*{Acknowledgements}
We thank Monash University and Hugging Face for providing compute instances. We are extremely grateful to Cristian Rojas for help on the initial exploration, Zhensu Sun for the discussion, Dmitry Abulkhanov for the paper review,  Brendan Dolan-Gavitt for providing the evaluation script of ``Asleep At The Keyboard'' benchmark, the BigCode community for providing the base models~\citep{li2023starcoder} and instruction tuning data~\citep{muennighoff2023octopack} from GitHub commits, and \cite{peft, hu2023llm} for implementing PEFT methods.

\bibliographystyle{acl_natbib}
\bibliography{reference}
\newpage
\appendix

\section{What is \textsc{Astraios}?}

\textsc{Astraios} is a suite of 28 instruction-tuned StarCoder models, employing 7 different PEFT methods across 4 model sizes, with up to 16B  parameters. Named after the Greek Titan god of the stars, \textsc{Astraios}, this model collection represents a vast array of ``stars'', each model illuminating a path to understanding the cost-performance trade-offs in Code LLMs. Through extensive testing across various tasks and datasets, \textsc{Astraios} evaluates the efficacy of fine-tuning methods with an emphasis on understanding their performance implications at different model scales, robustness, and security aspects. The suite serves as a celestial guide in the Code LLM universe, helping to chart the most efficient and effective methods for model fine-tuning.

\section{Artifacts}

\begin{table*}[h]
    \centering
    \resizebox{1\textwidth}{!}{
    \begin{tabular}{l|c}
    \toprule
    Name & Public Link\\
    \midrule
    \multicolumn{2}{c}{\textit{Base Models}} \\
    \midrule
    StarCoderBase 1B  & \url{https://huggingface.co/bigcode/starcoderbase-1b} \\
    StarCoderBase 3B  & \url{https://huggingface.co/bigcode/starcoderbase-3b} \\    
    StarCoderBase 7B  & \url{https://huggingface.co/bigcode/starcoderbase-7b} \\    
    StarCoderBase  & \url{https://huggingface.co/bigcode/starcoderbase} \\    
    \midrule
    
    \multicolumn{2}{c}{\textit{Instruction Tuning Data}} \\
    \midrule
    CommitPackFT + OASST & \url{https://huggingface.co/datasets/bigcode/guanaco-commits} \\
    \midrule
    \multicolumn{2}{c}{\textit{Original PEFT Implementation}} \\
    \midrule
    LoRA & \url{https://github.com/huggingface/peft} \\
    P-Tuning & \url{https://github.com/huggingface/peft} \\
    Adapter$^H$ & \url{https://github.com/AGI-Edgerunners/LLM-Adapters} \\
    Adapter$^P$ & \url{https://github.com/AGI-Edgerunners/LLM-Adapters} \\
    Parallel & \url{https://github.com/AGI-Edgerunners/LLM-Adapters} \\
    (IA)$^3$ & \url{https://github.com/huggingface/peft} \\
    Prompt & \url{https://github.com/huggingface/peft} \\
    AdaLoRA & \url{https://github.com/huggingface/peft} \\
    \midrule
    \multicolumn{2}{c}{\textit{Evaluation Framework}} \\
    \midrule
    Code Generation LM Evaluation Harness & \url{https://github.com/bigcode-project/bigcode-evaluation-harness} \\
    \midrule
    \multicolumn{2}{c}{\textit{Astraios Models}} \\
    \midrule
    Astraios LoRA 1B & \url{https://huggingface.co/bigcode/astraios-1b-lora}\\
    Astraios P-Tuning 1B & \url{https://huggingface.co/bigcode/astraios-1b-ptuning}\\
    Astraios Adapter$^H$ 1B & \url{https://huggingface.co/bigcode/astraios-1b-adapterh}\\
    Astraios Adapter$^P$ 1B & \url{https://huggingface.co/bigcode/astraios-1b-adapterp}\\
    Astraios Parallel 1B & \url{https://huggingface.co/bigcode/astraios-1b-parallel}\\
    Astraios (IA)$^3$  1B & \url{https://huggingface.co/bigcode/astraios-1b-ia3}\\
    Astraios LoRA 3B & \url{https://huggingface.co/bigcode/astraios-3b-lora}\\
    Astraios P-Tuning 3B & \url{https://huggingface.co/bigcode/astraios-3b-ptuning}\\
    Astraios Adapter$^H$ 3B & \url{https://huggingface.co/bigcode/astraios-3b-adapterh}\\
    Astraios Adapter$^P$ 3B & \url{https://huggingface.co/bigcode/astraios-3b-adapterp}\\
    Astraios Parallel 3B & \url{https://huggingface.co/bigcode/astraios-3b-parallel}\\
    Astraios (IA)$^3$  3B & \url{https://huggingface.co/bigcode/astraios-3b-ia3}\\
    Astraios LoRA 7B & \url{https://huggingface.co/bigcode/astraios-7b-lora}\\
    Astraios P-Tuning 7B & \url{https://huggingface.co/bigcode/astraios-7b-ptuning}\\
    Astraios Adapter$^H$ 7B & \url{https://huggingface.co/bigcode/astraios-7b-adapterh}\\
    Astraios Adapter$^P$ 7B&  \url{https://huggingface.co/bigcode/astraios-7b-adapterp}\\
    Astraios Parallel 7B & \url{https://huggingface.co/bigcode/astraios-7b-parallel}\\
    Astraios (IA)$^3$  7B & \url{https://huggingface.co/bigcode/astraios-7b-ia3}\\
    Astraios LoRA 16B & \url{https://huggingface.co/bigcode/astraios-lora}\\
    Astraios P-Tuning 16B & \url{https://huggingface.co/bigcode/astraios-ptuning}\\
    Astraios Adapter$^H$ 16B & \url{https://huggingface.co/bigcode/astraios-adapterh}\\
    Astraios Adapter$^P$ 16B & \url{https://huggingface.co/bigcode/astraios-adapterp}\\
    Astraios Parallel 16B & \url{https://huggingface.co/bigcode/astraios-parallel}\\
    Astraios (IA)$^3$  16B & \url{https://huggingface.co/bigcode/astraios-ia3}\\
    \bottomrule
    \end{tabular}
    }
    \caption{
        \textbf{Used and produced artifacts.}
    }
    \label{tab:artifacts}
\end{table*}
\newpage

\section{Contributions}
Terry Yue Zhuo trained 1B and 3B models, conducted most evaluations, analyzed the results, wrote the paper and led the project. Armel Zebaze trained 7B and 15B models, evaluated 15B models on Code Synthesis and Code Repair, analyzed the results and helped edit the paper. Nitchakarn Suppattarachai evaluated two comprehension tasks. Niklas Muennighoff advised on the experiments and helped with plotting. Niklas Muennighoff, Qian Liu, Harm de Vries and Leandro von Werra provided suggestions and helped edit the paper.

\section{Instruction Tuning}
All the instruction tuning experiments have been conducted on A100 80G GPUs.
For all PEFT strategies, we use the 8-bit quantized base models for training. For FFT, we use the original base models without quantization.

\label{app:peft}
\paragraph{LoRA} We use the attention dimension of 8, the alpha parameter of 16, dropout probability of 0.05, and target modules of "[c\_proj, c\_attn, q\_attn]". We keep the other hyperparameters as default.
\paragraph{P-Tuning} We use the 30 virtual tokens and remain the other hyperparameters as default.
\paragraph{Adapter$^H$} We use target modules of "[c\_fc, mlp.c\_proj]". We keep the other hyperparameters as default.
\paragraph{Adapter$^P$} We use target modules of " [mlp.c\_proj]". We keep the other hyperparameters as default.
\paragraph{Parallel} We use target modules of "[c\_fc, mlp.c\_proj]". We keep the other hyperparameters as default.
\paragraph{(IA)$^3$} We target modules of "c\_attn, mlp.c\_proj]" and feedforward modules of " [mlp.c\_proj]".
\paragraph{Prompt~\citep{lester2021power}} We use the 30 virtual tokens and keep the other hyperparameters as default.
\paragraph{AdaLoRA~\citep{zhang2022adaptive}} We use the target average rank of the incremental matrix of 8, the initial rank for each incremental matrix of 12, 200 steps of initial fine-tuning warmup, 1000 step of final fine-tuning, the alpha parameter of 16, dropout probability of 0.05, the time interval between two budget allocations of 10, EMA for sensitivity smoothing of 0.85, EMA for uncertainty quantification of 0.85, and target modules of "[c\_proj, c\_attn, q\_attn]". We keep the other hyperparameters as default.

\section{Evaluation Setup}

\paragraph{Devign} We generate the outputs with a max length of 512 tokens in the style of greedy decoding. All other parameters are defaulted in \cite{bigcode-evaluation-harness}. For the one-shot example, we randomly sample from the train set.

\paragraph{BigCloneBench} We generate the outputs with a max length of 512 tokens in the style of greedy decoding. All other parameters are defaulted in \cite{bigcode-evaluation-harness}. For the one-shot example, we randomly sample from the train set.

\paragraph{HumanEvalPack} We generate 20 outputs per example with a max length of 2048 tokens and a temperature of 0.2. All other parameters are defaulted in \cite{bigcode-evaluation-harness}.

\paragraph{ReCode} We generate the outputs with a max length of 1024 tokens in the style of greedy decoding. All other parameters are defaulted in \cite{bigcode-evaluation-harness}.

\paragraph{Asleep At The Keyboard} We generate 20 outputs per example with a max length of 1024 tokens and a temperature of 0.2. All other parameters are defaulted in \cite{bigcode-evaluation-harness}.

\section{Failure of Scaling}
\label{app:fail}

\begin{figure}[H]
\centering
\begin{minipage}[b]{0.45\linewidth}
\includegraphics[width=\linewidth]{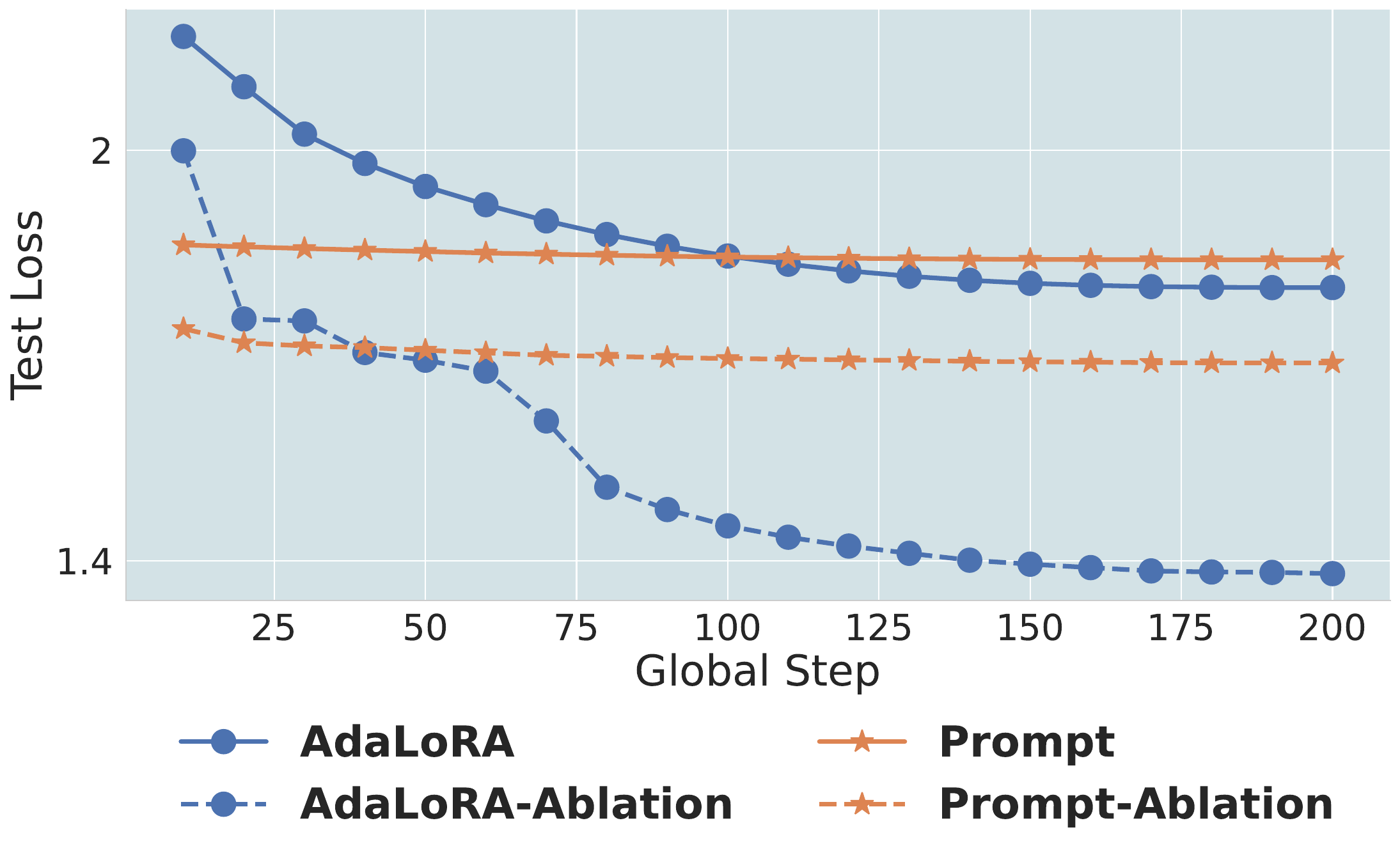}
\caption*{1B model.}
\end{minipage}
\hfill
\begin{minipage}[b]{0.45\linewidth}
\includegraphics[width=\linewidth]{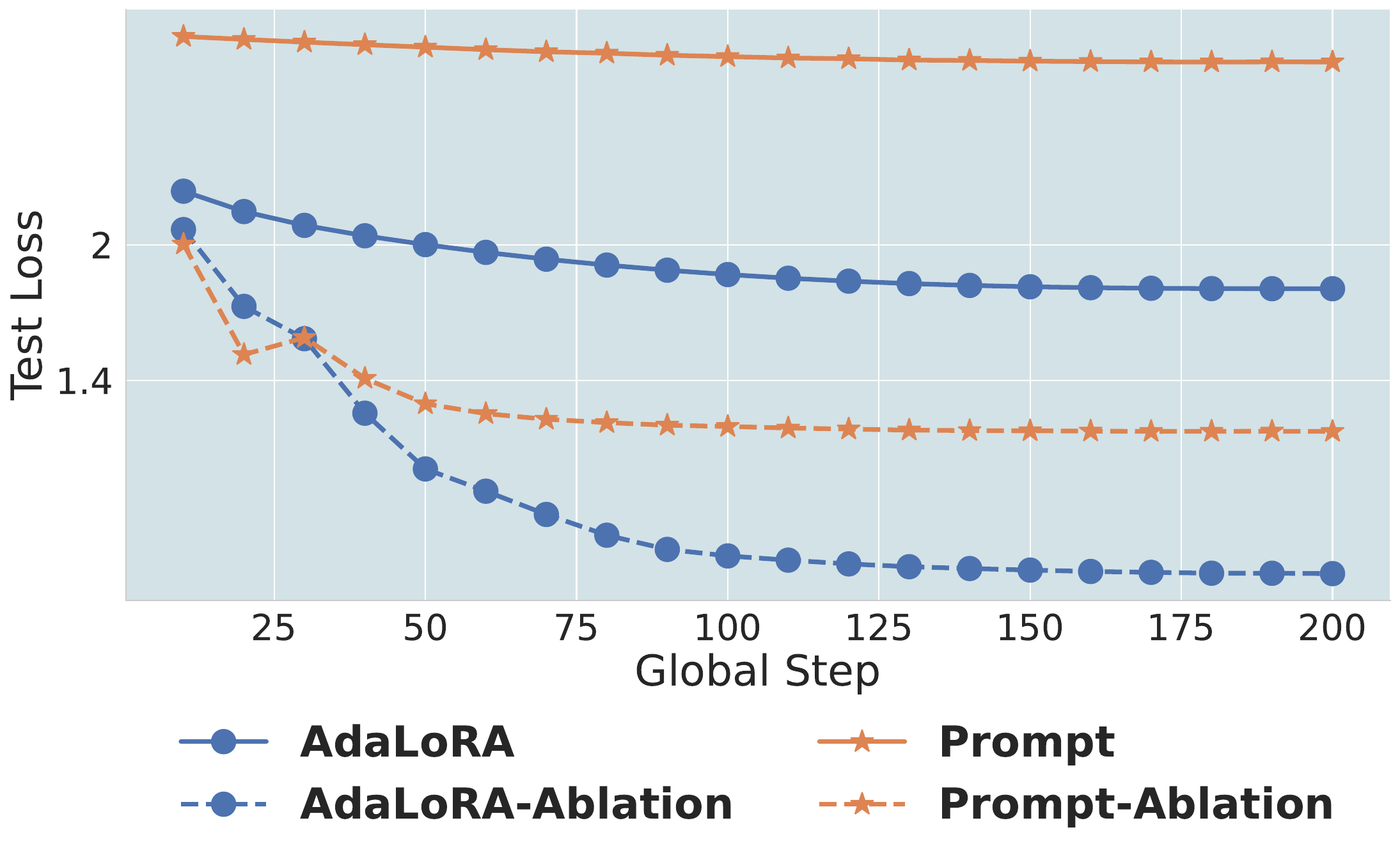}
\caption*{3B models.}
\end{minipage}
\caption{Test loss of selected models across training time measured by \textit{Global Step}.}
\label{fig:ablation_training_time}
\end{figure}

\begin{figure}[H]
\centering
\begin{minipage}[b]{0.45\linewidth}
\includegraphics[width=\linewidth]{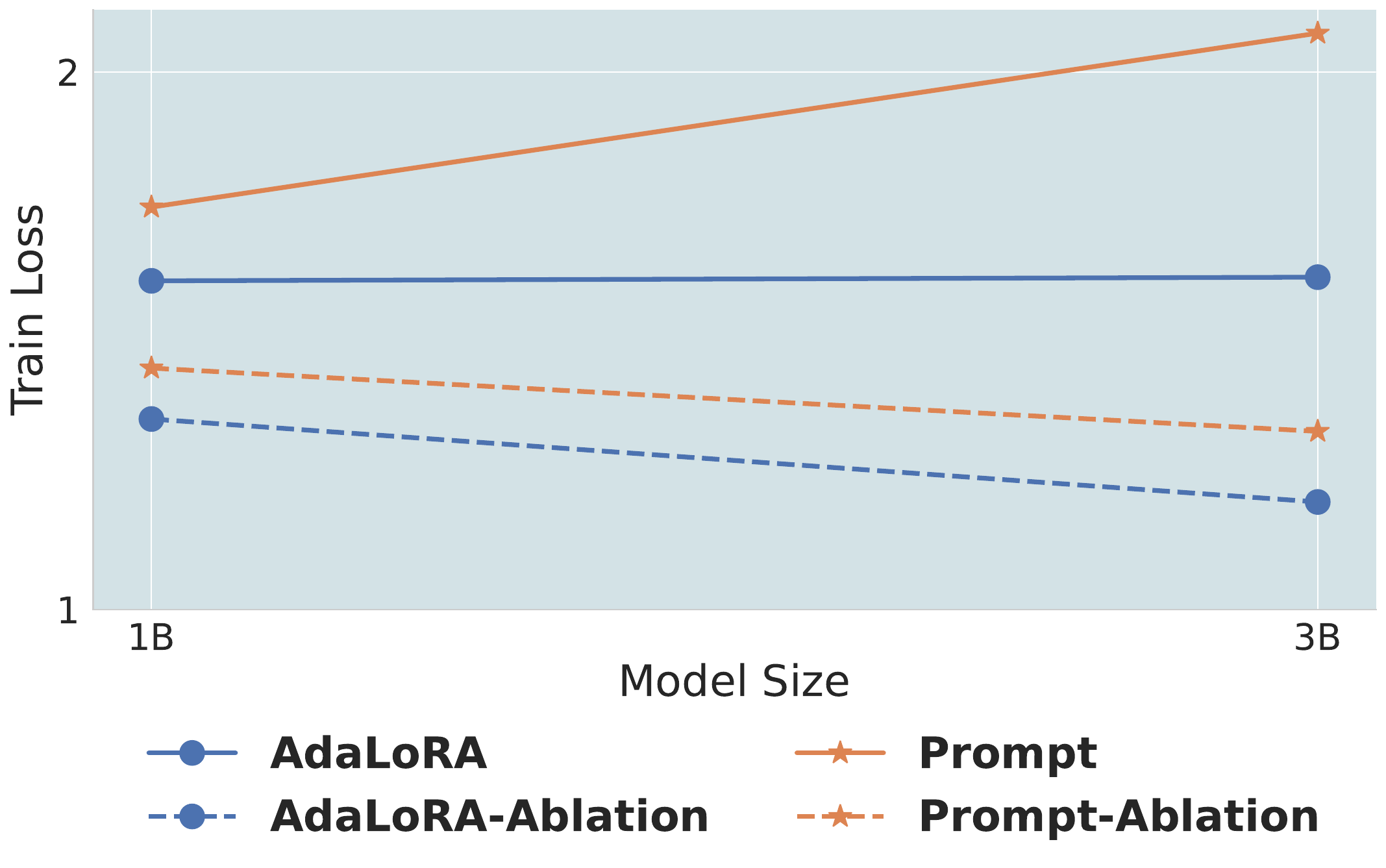}
\end{minipage}
\hfill
\begin{minipage}[b]{0.45\linewidth}
\includegraphics[width=\linewidth]{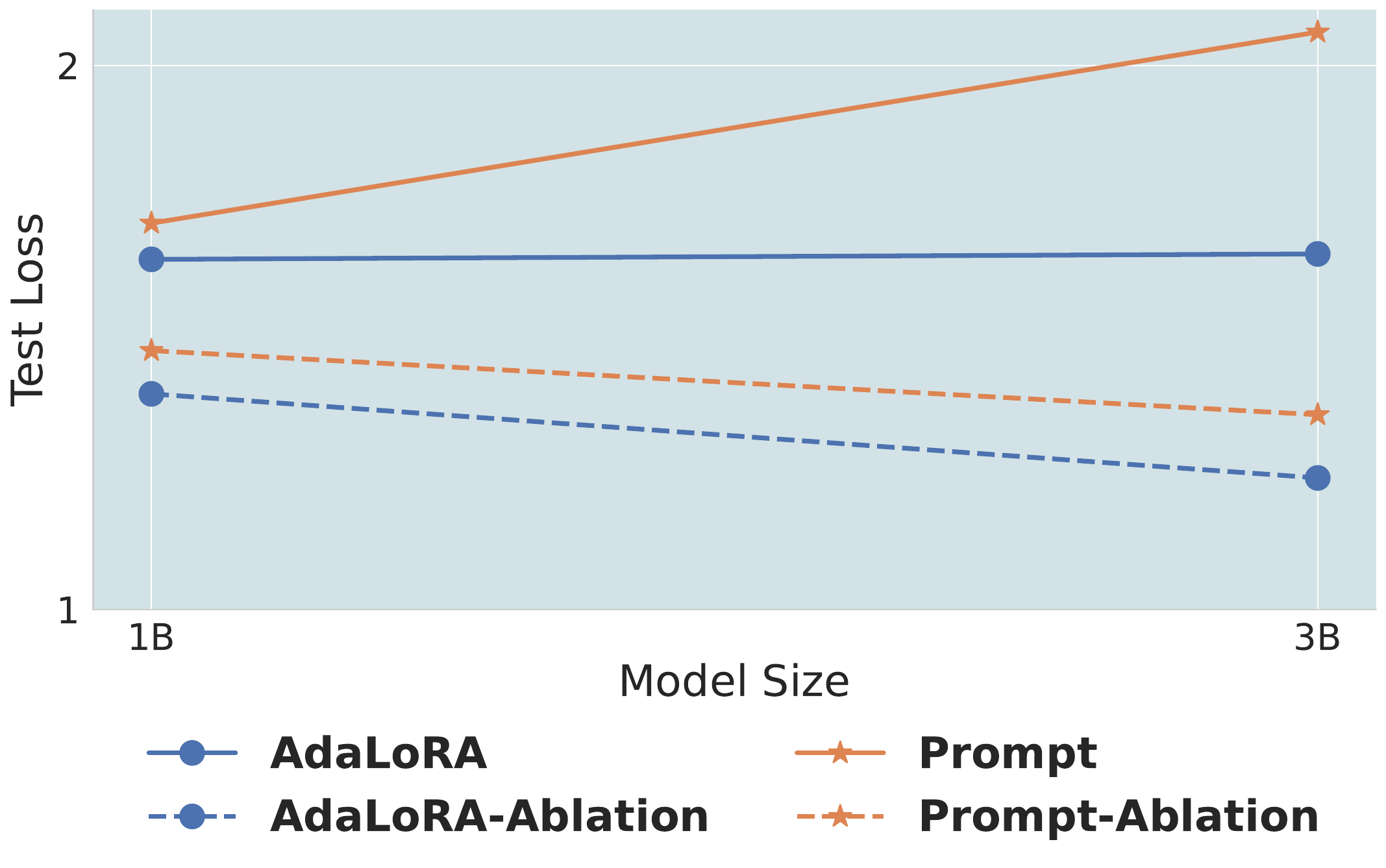}
\end{minipage}
\caption{Final loss across model sizes.}
\label{fig:ablation_model_size}
\end{figure}

During the initial experiment, we also train the models with Prompt Tuning~\citep{lester2021power} and AdaLoRA~\citep{zhang2022adaptive}. Although the loss continues decreasing when the training time increases, we observe the phenomenon of model size scales in contrast to Section~\ref{sec:training}. As shown in Figure~\ref{fig:ablation_model_size}, the final loss of these two tuning strategies consistently increases as the model size increases, which is contrary to what we observe for other PEFT methods. In the new version of LLM-Adapter~\citep{hu2023llm}, we notice that the learning rate has been specifically mentioned. For Prompt Tuning, the authors use $3 \times 10^{-2}$ instead of $3 \times 10^{-4}$, which is used in their other selected PEFT strategies. Therefore, we hypothesize that some tuning strategies may require a much higher learning rate to achieve optimal performance. We further try a few learning rates on training 1B and 3B StarCoderBase models and find that $3 \times 10^{-2}$ works well for Prompt Tuning. In addition, $3 \times 10^{-2}$ and $1 \times 10^{-3}$ also work much better for AdaLoRA. With the new set of learning rates, we find that these tuning strategies are aligned with our findings in Section~\ref{sec:loss}. Different from the conclusion of \cite{kaplan2020scaling} that the choice of learning rate schedule is mostly irrelevant in language model pre-training, we suggest that hyperparameters of learning rate schedule may matter a lot for scaling parameter-efficient language model on fine-tuning.

\section{Visualization on HumanEvalPack}

\begin{figure}[H]
\centering

\begin{minipage}[b]{0.45\linewidth}
\includegraphics[width=\linewidth]{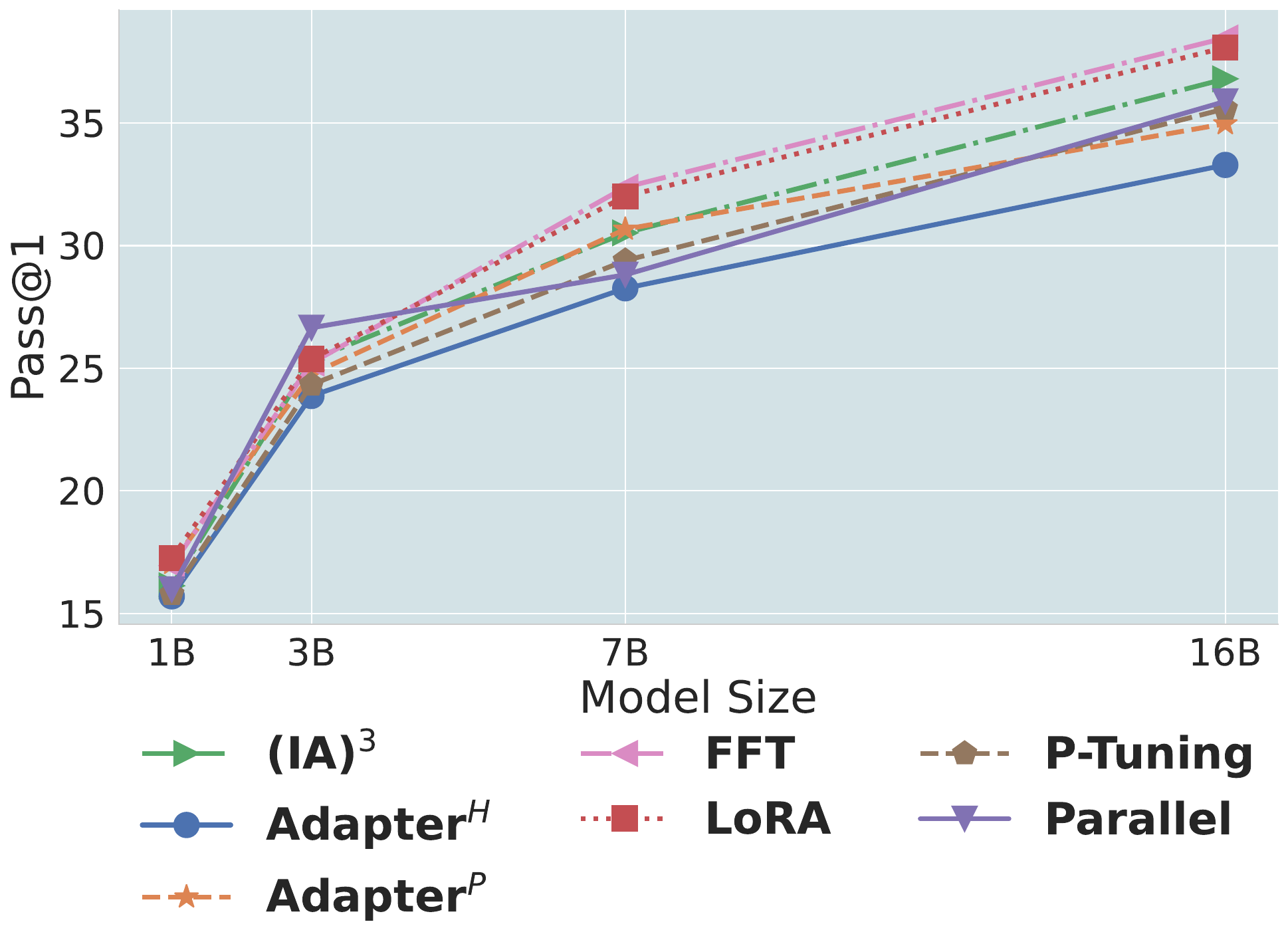}
\caption*{Python Code Synthesize}
\end{minipage}
\hfill
\begin{minipage}[b]{0.45\linewidth}
\includegraphics[width=\linewidth]{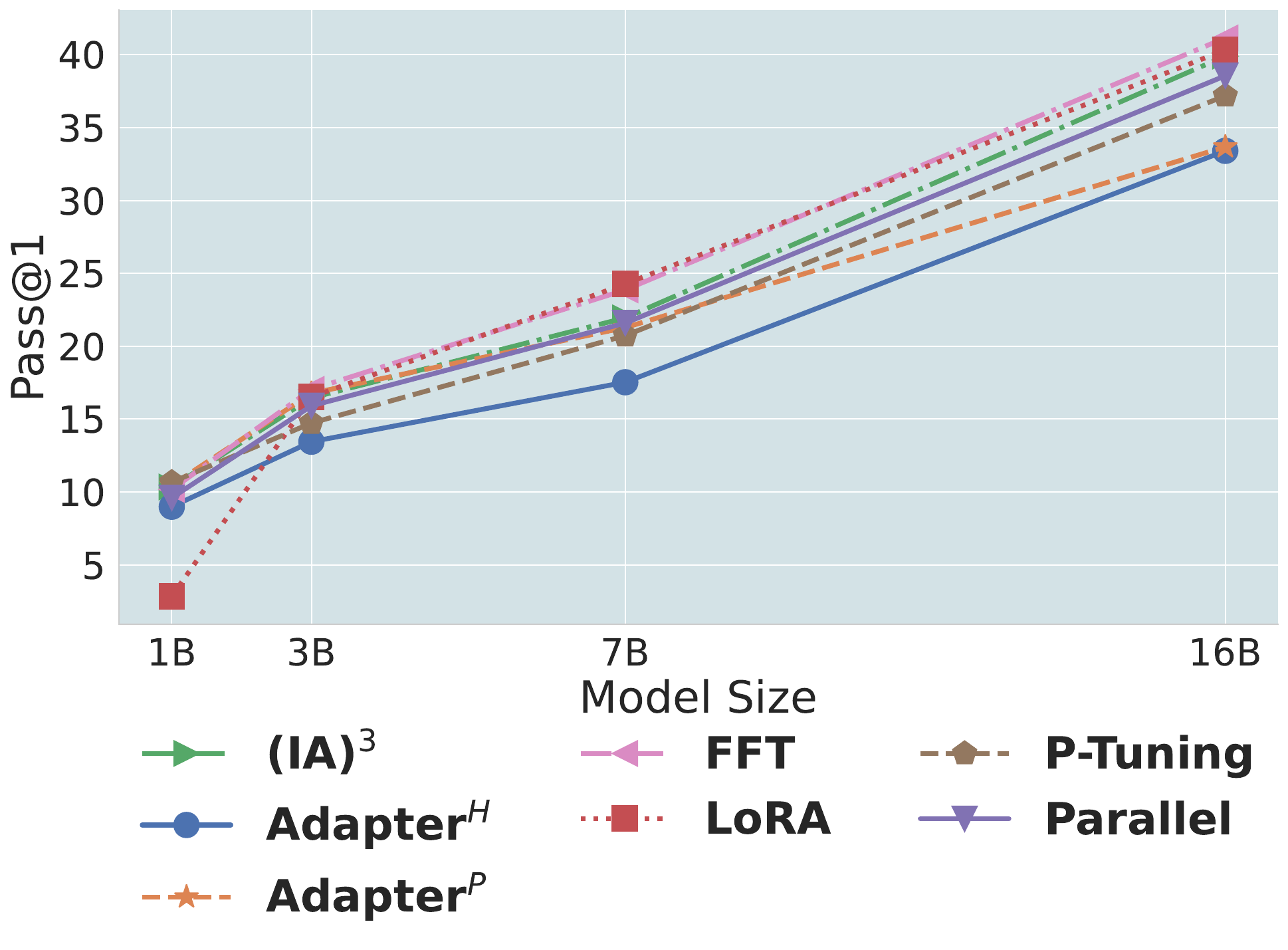}
\caption*{Java Code Synthesize}
\end{minipage}
\vspace{1em}

\begin{minipage}[b]{0.45\linewidth}
\includegraphics[width=\linewidth]{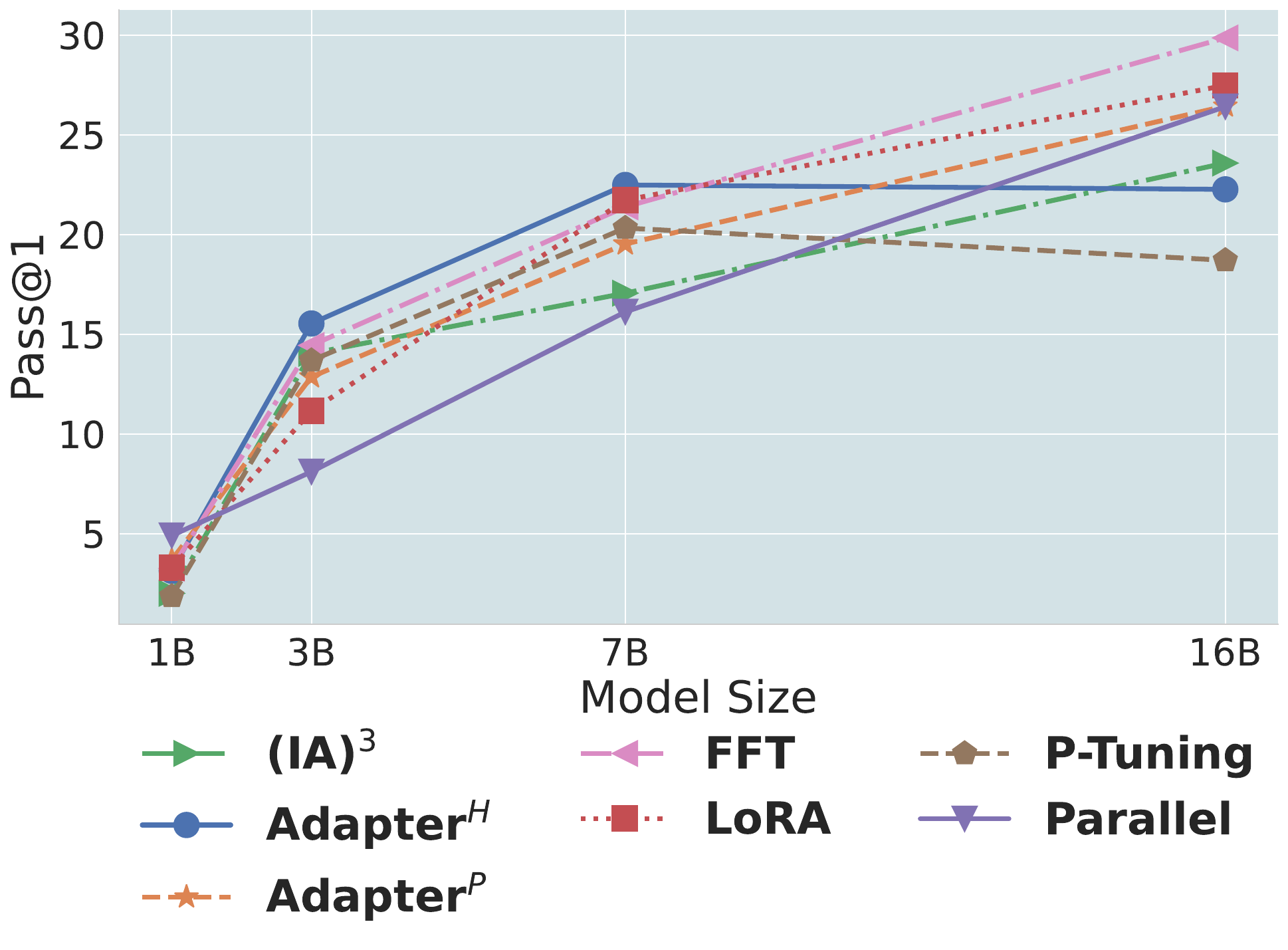}
\caption*{Python Code Repair}
\end{minipage}
\hfill
\begin{minipage}[b]{0.45\linewidth}
\includegraphics[width=\linewidth]{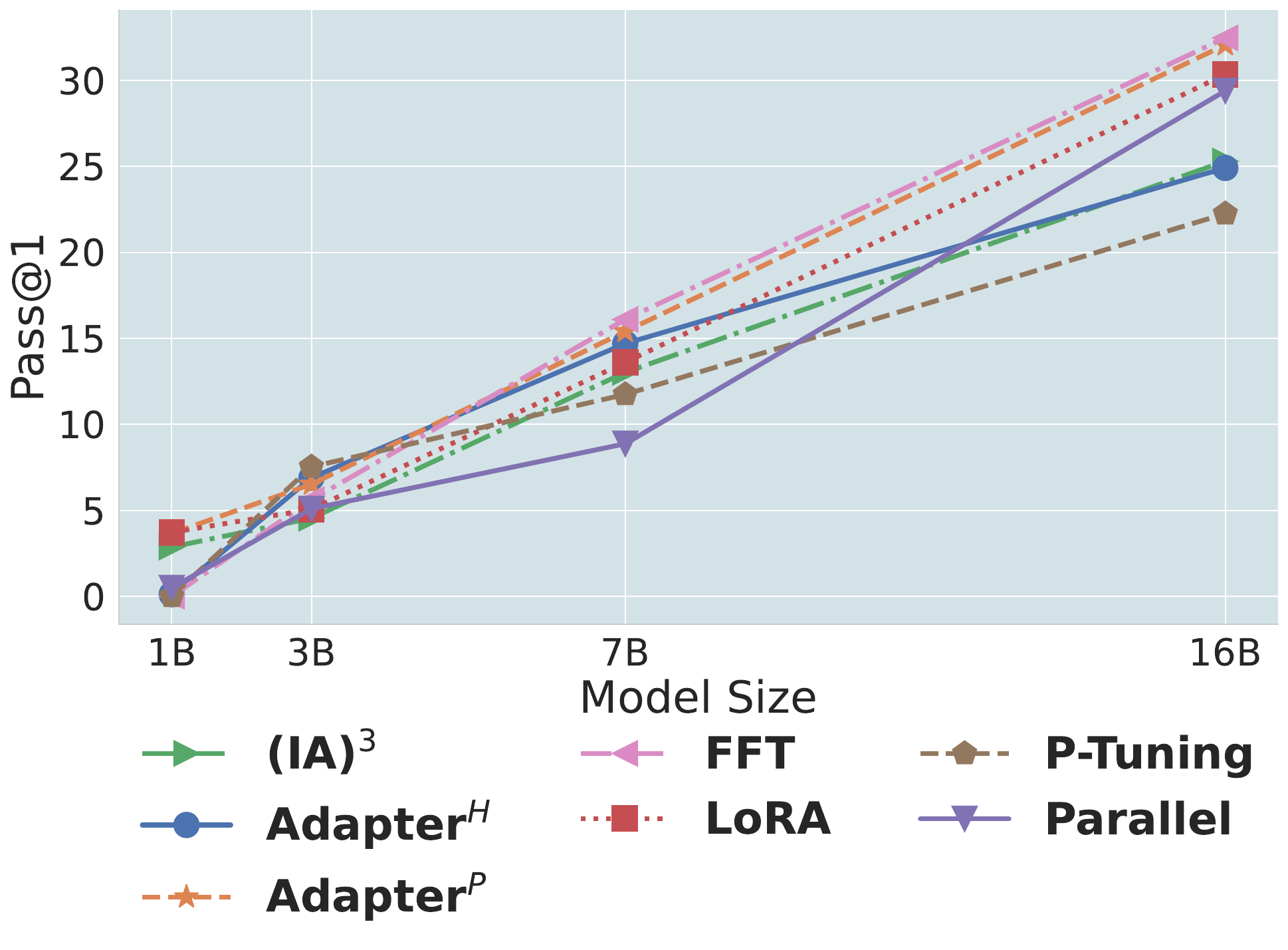}
\caption*{Java Code Repair}
\end{minipage}
\vspace{1em}

\begin{minipage}[b]{0.45\linewidth}
\includegraphics[width=\linewidth]{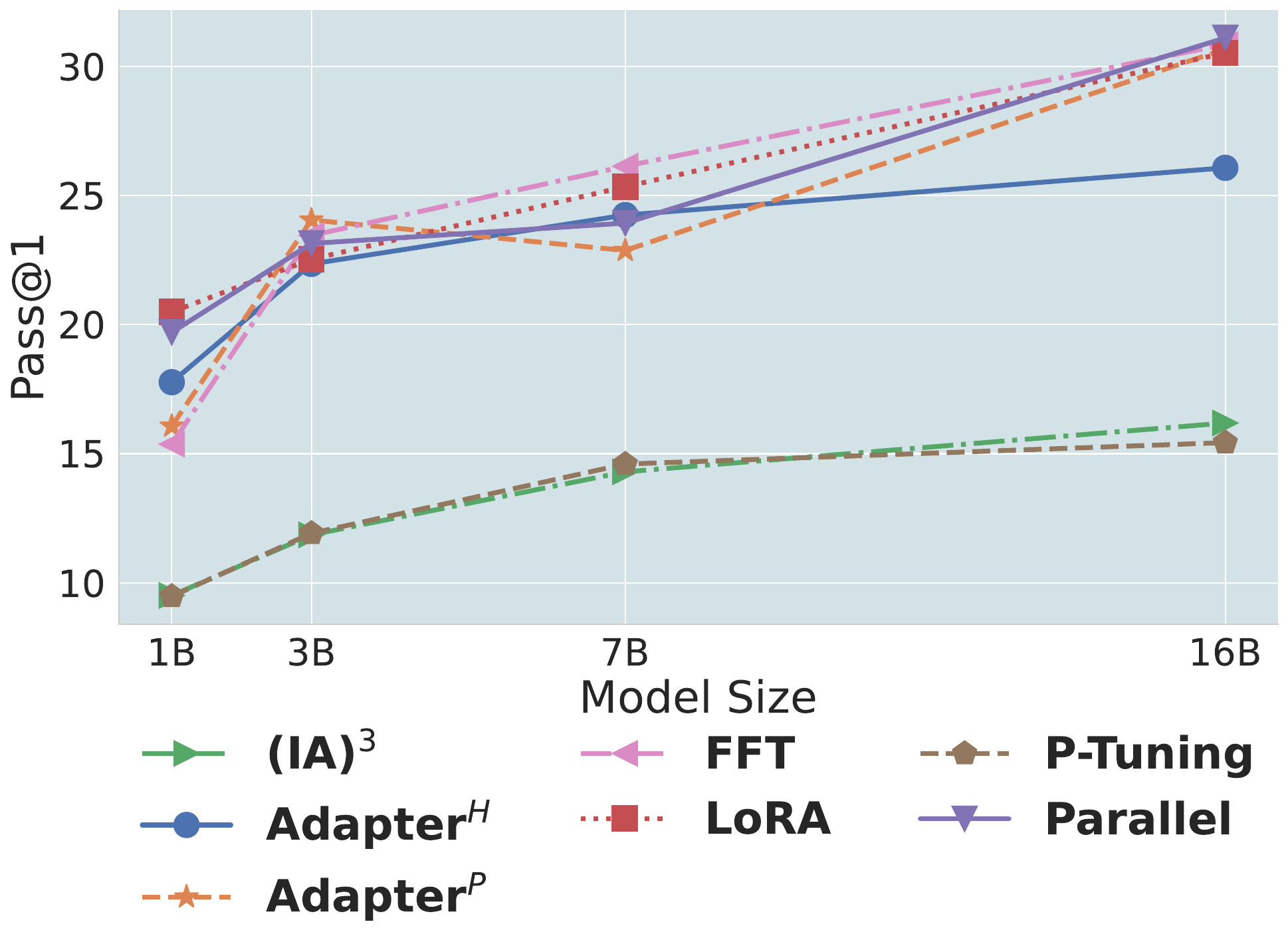}
\caption*{Python Code Explain}
\end{minipage}
\hfill
\begin{minipage}[b]{0.45\linewidth}
\includegraphics[width=\linewidth]{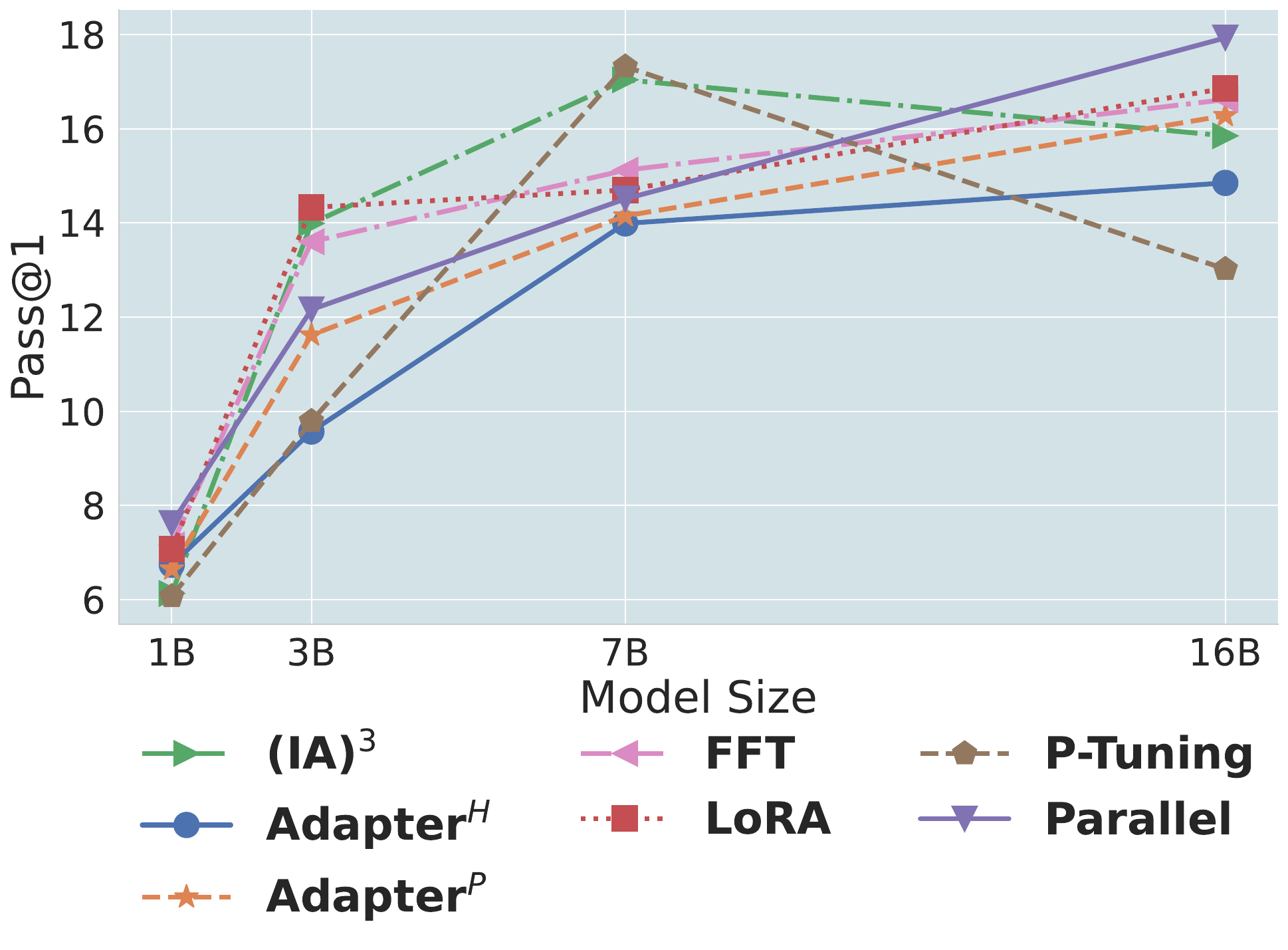}
\caption*{Java Code Explain}
\end{minipage}

\caption{Pass@1 results of \textsc{Astraios} models on HumanEvalPack.}
\label{fig:figures}

\end{figure}

\section{Mitigating Inverse Scaling} 
\begin{figure}[H]
\centering

\begin{minipage}[b]{0.45\linewidth}
\includegraphics[width=\linewidth]{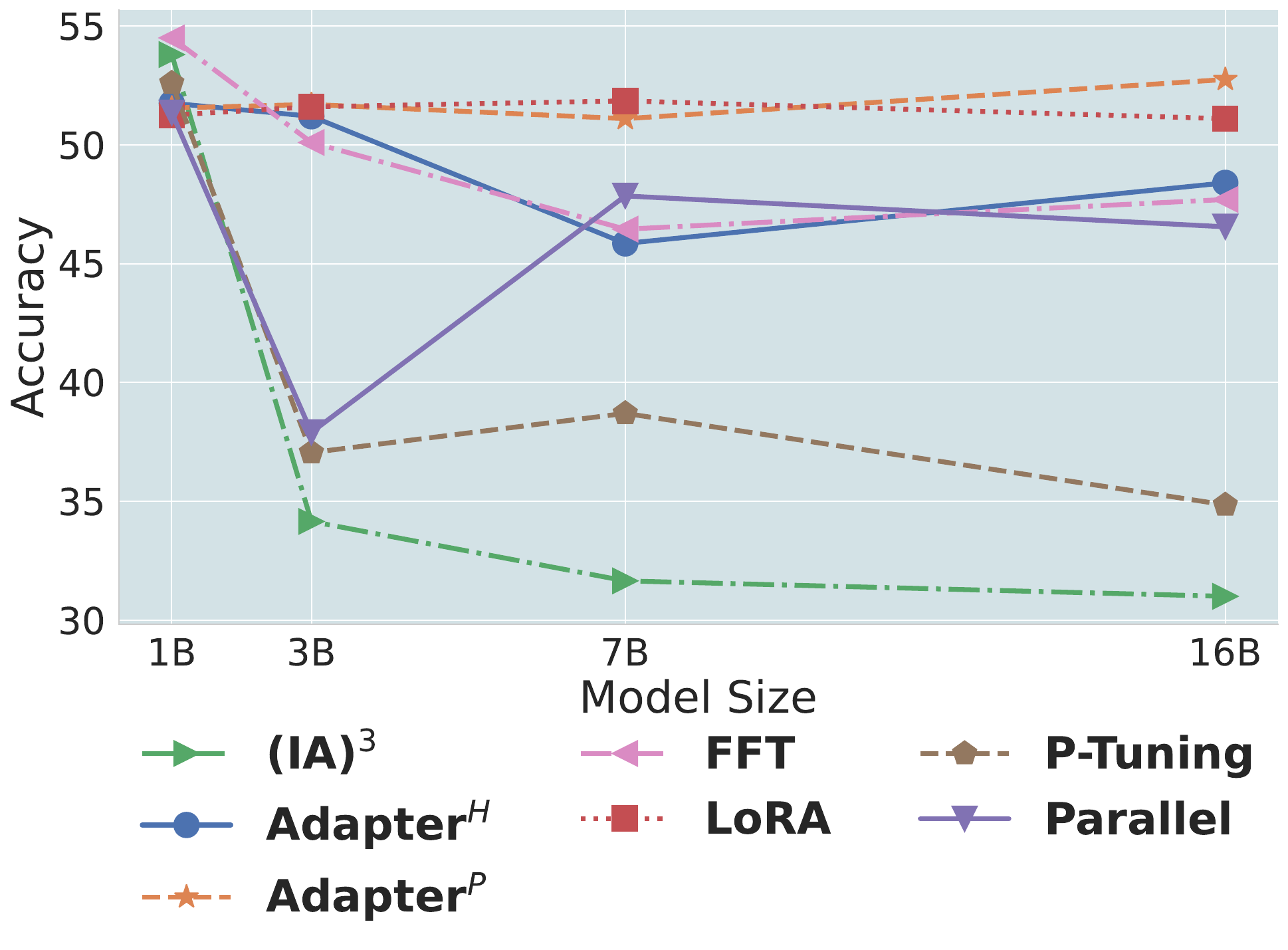}
\caption{Results on Defect Detection with 1-shot demonstration.}
\label{fig:ablation_defect_defection}
\end{minipage}
\hfill
\begin{minipage}[b]{0.45\linewidth}
\includegraphics[width=\linewidth]{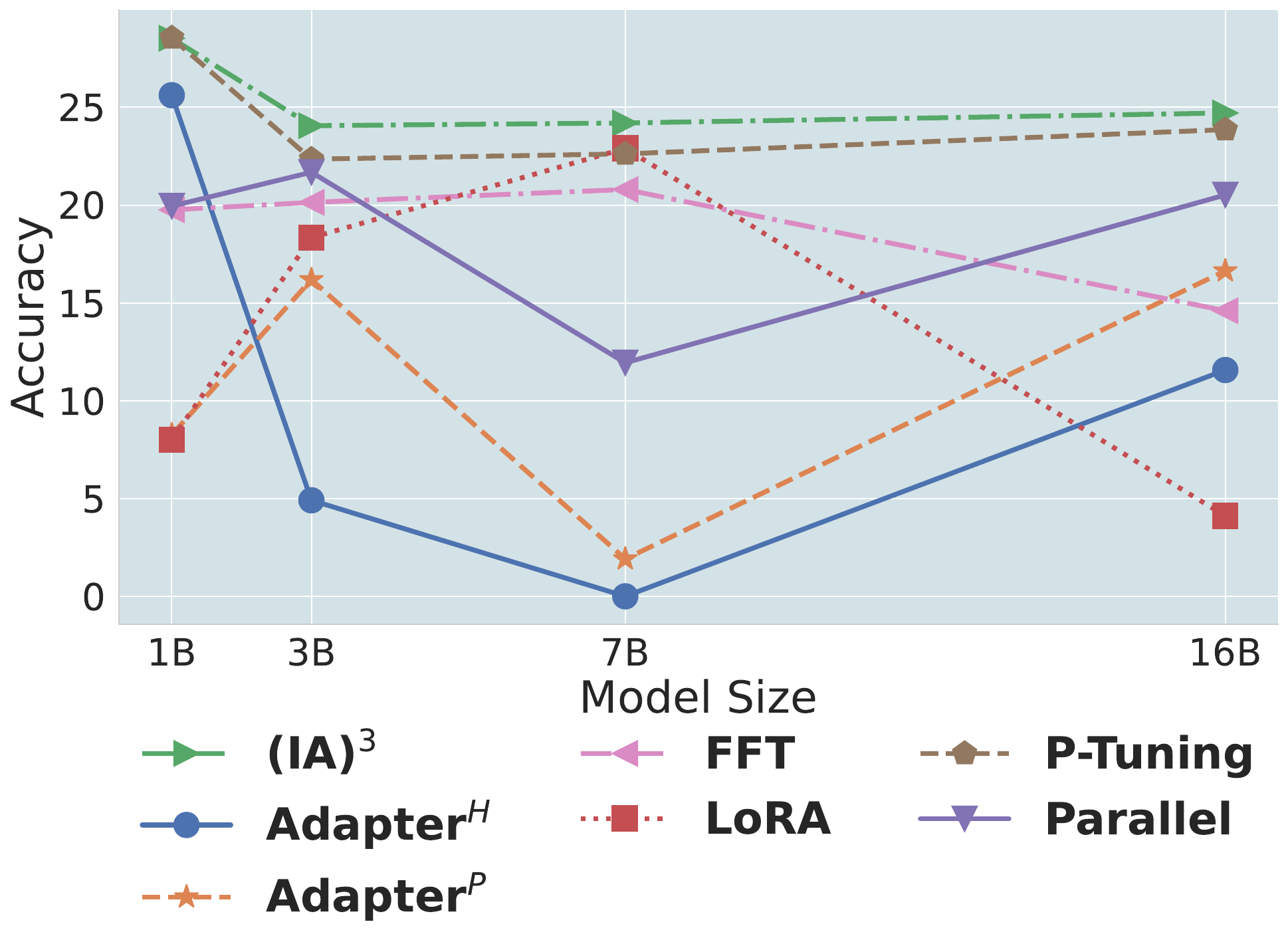}
\caption{Results on Clone Detection with 1-shot demonstration.}
\label{fig:ablation_clone_defection}
\end{minipage}
\end{figure}
We have attempted to see if the inverse-scaling-like patterns in code comprehension tasks can be mitigated and more aligned with scaling laws. As \cite{wei2022inverse} have shown that 1-shot demonstrations can make all inverse scaling tasks U-shaped or flat, we try to see if 1-shot examples can help with defection detection and clone detection. To select the 1-shot examples, we randomly sample a fixed sample from the train set of each benchmark. We re-evaluate all \textsc{Astraios} models on the two tasks and present the results in Figures~\ref{fig:ablation_defect_defection} and \ref{fig:ablation_clone_defection}. For defect detection, all PEFT strategies become flatter than the previous patterns, which is similar to what \cite{wei2022inverse} observe. However, for clone detection, the patterns of some tuning strategies like LoRA and FFT do not turn flat. Although the performances of LoRA and FFT have been scaling up to 7B, they decrease at 15B. We hypothesize that our size scaling is still not significant enough to represent an increasing pattern after 15B for LoRA and FFT with 1-shot demonstrations. 

\section{Further Discussion}
We further measure the correlations among final loss in Section~\ref{sec:loss}, overall task performance in Section~\ref{sec:task}, and numbers of updated parameters via three metrics, Kendall ($\tau$), Pearson ($r_p$), and Spearman ($r_s$) coefficients. Kendall coefficient measures the ordinal association and is robust against outliers, making it useful for non-normal data distributions. Pearson's coefficient assesses linear correlation, which is ideal for normal data distributions with expected linear relationships. Spearman's coefficient, like Kendall coefficient, is a non-parametric measure that assesses rank correlation, useful for identifying monotonic but non-linear relationships.
\begin{table}[h]
\centering
\caption{Correlations between trainable parameters and final loss. Correlations are computed with the negated loss ($-$loss), so a positive coefficient means that more trainable parameters are associated with lower loss. $p$-values are provided in \textcolor{gray}{gray}.}
\label{tab:para_loss}
\resizebox{0.8\textwidth}{!}{
\begin{tabular}{c|ccc|ccc}
\toprule
\multirow{2}{*}{\textbf{Model Size}} & \multicolumn{3}{c|}{\textbf{Train Loss}} & \multicolumn{3}{c}{\textbf{Test Loss}} \\
\cmidrule{2-7}
 & $\tau$ & $r_p$ & $r_s$ & $\tau$ & $r_p$ & $r_s$ \\
\midrule
1B 
& .4286
 & .3113
 & .6071
  & .3333
 & .3358
 & .4643
 \\
3B 
 & .5238
 & .3433
 & .7143
  & .2381
 & .3835
 & .4286
\\
7B 
 & .5238
 & .3555
 & .7143
 & .2381
 & .4091
 & .4286
\\
16B 
 & .5238
 & .3524
 & .7143
 & .2381
 & .3986
 & .4286
\\\midrule
Overall
 & .4339 \textcolor{gray}{(.00)}
 & .3328 \textcolor{gray}{(.08)}
 & .5616 \textcolor{gray}{(.00)}
 & .3598 \textcolor{gray}{(.01)}
 & .3308 \textcolor{gray}{(.09)}
 & .4953 \textcolor{gray}{(.01)}
\\
\bottomrule
\end{tabular}}
\end{table}

We compute the correlations between updated parameters of \textsc{Astraios} models and the final loss of corresponding models in Table~\ref{tab:para_loss}. From the table, we first observe that the updated parameters are more correlated to the final train loss than the test loss. However, they all imply that there is a moderated correlation, which can be used for cross-entropy loss in model training. We also observe that when we aggregate all statistics across model sizes, the correlations may slightly decrease.

\begin{table}[h]
\centering
\caption{Correlations between final loss and overall task performance. Correlations are computed with the negated loss ($-$loss), so a positive coefficient means that lower loss is associated with better task performance. $p$-values are provided in \textcolor{gray}{gray}.}
\label{tab:loss_downstream}
\resizebox{0.8\textwidth}{!}{
\begin{tabular}{c|ccc|ccc}
\toprule
\multirow{2}{*}{\textbf{Model Size}} & \multicolumn{3}{c|}{\textbf{Train Loss}} & \multicolumn{3}{c}{\textbf{Test Loss}}\\
\cmidrule{2-7}
 & $\tau$ & $r_p$ & $r_s$ & $\tau$ & $r_p$ & $r_s$ \\
\midrule
1B 
& -.2381
 & -.4319
 & -.2857
 & -.0476
 & -.4328
 & -.0357
 \\
3B 
& .5238
 & .7819
 & .7143
& .8095
 & .7859
 & .9286

\\
7B 
 & .5238
 & .7165
 & .6786
 & .8095
 & .8230
 & .9286

\\
16B 
& .3333
 & .8096
 & .5000
 & .8095
 & .9211
 & .8929

\\\midrule
Overall
 & .7302 \textcolor{gray}{(.00)}
 & .9027 \textcolor{gray}{(.00)}
 & .9201 \textcolor{gray}{(.00)}
 & .8466 \textcolor{gray}{(.00)}
 & .9277 \textcolor{gray}{(.00)}
 & .9579 \textcolor{gray}{(.00)}
\\
\bottomrule
\end{tabular}}
\end{table}

We compute the correlations between the model loss and their mean downstream scores calculated in Section~\ref{sec:task}. We show the results in Table~\ref{tab:loss_downstream}, where we compute correlations for each model size and the final aggregated statistics. Our observation on the size-level correlations indicates that the task performance of 1B models is hard to align with the final loss, while bigger models tend to be much more correlated to both train and test loss. We explain the hypothesis that 1B models do not have enough capability to learn instructions. When aggregating the data points, we find that correlations are much stronger than the size-level prediction. The strong correlations imply that model loss on the general instruction data can work as a good proxy of downstream tasks in Code LLMs. When comparing the correlations on train loss to the test loss, we observe the correlations are stronger on the latter one. This can be explained by the fact that models tend to FFT on the training data, where the loss on the train split can not generalize well on the unseen tasks and data.  Moreover, we also ask: \textit{What is the relationship between the downstream task performance and the updated parameters?} Therefore, We investigate the correlation between tuned parameters and cumulative scores. The correlations are 0.3016 \textcolor{gray}{(.02)}, 0.4128 \textcolor{gray}{(.03)}
and 0.4138 \textcolor{gray}{(.03)} for Kendall, Pearson and Spearman correlations, respectively. We draw the conclusion -- \textit{Possible}.

\section{Limitations and Future Work}

\paragraph{Experiment Noise} We observe that our empirical results are based solely on a single run of each task, due to budget constraints that prevent us from tuning and evaluating the same Code LLMs multiple times. Although the single evaluation approach limits the breadth of our results and may introduce unexpected experiment noise, it provides a preliminary insight into the performance and potential of PEFT in different scenarios. Future investigations with multiple runs are necessary to establish more robust conclusions and understand the variance and reliability of our results.

\paragraph{Fair Evaluation} To compare different PEFT strategies fairly, we have used the same training configurations described in Section~\ref{sec:training}. However, as we find that some PEFT strategies like Prompt Tuning may be sensitive to the training hyperparameters in Section~\ref{sec:loss}, the consistent configurations can be unfair. On the other hand, finding the optimal hyperparameters for each PEFT strategy is impractical and can cost more than training with FFT. A more efficient approach is to reuse the hyperparameters in previous work, which motivates us to adopt the default settings in the PEFT library and LLM-Adapter framework. Meanwhile, we believe there may be other practical approaches to benchmark PEFT strategies, encouraging the community to investigate further.

\paragraph{PEFT Strategy} We notice that there are many more PEFT strategies~\citep{karimi2021compacter,zaken2022bitfit,wang2022multitask,edalati2022krona} have been proposed recently. Due to the limited computation budget, we do not include them all in our \textsc{Astraios} model suite. However, we have publicly made all our source code, data, and models available. We encourage future development in analyzing PEFT strategies on Code LLMs, which helps design more efficient PEFT strategies.

\paragraph{Data Scaling} One limitation of our work is that we do not verify the validity of data scaling on PEFT strategies. However, this factor has been well-studied in various works~\citep {kaplan2020scaling,hoffmann2022training,muennighoff2023scaling} for model pre-training and fine-tuning. As we find that the performance of PEFT on Code LLMs monotonically increases when scaling up the model size and training time, these selected PEFT strategies are likely aligned with the previous findings of data scaling. We recommend further verification on this aspect.

\paragraph{Model Architecture} Another limitation of our study is that we do not vary the model architecture of Code LLMs. It is possible that some findings may not generalize to other encoder-decoder Code LLMs like CodeT5~\citep{wang2021codet5} and CodeT5+~\citep{wang2023codet5+}. However, as StarCoder is built upon the enhanced GPT-2~\citep{radford2019language} architecture, we believe that our observations can be transferred to other GPT-based LLMs.

\paragraph{Scaling Parameter-Constrained Language Models} Although we demonstrate the possibility of predicting the final loss based on the updated parameters and vice versa, we note that a scaling law generally needs more than 100 models and their final loss. Ideally, the training experiments should be consistent with different PEFT strategies, meaning that training hundreds of models is needed. Furthermore, task performance is hard to predict, as there is much more noise in the downstream tasks than the final loss. We foresee that predicting such overall performance is very challenging.

\section{Prompts}

The prompting format can significantly impact performance. In the spirit of true few-shot learning~\citep{perez2021true}, we do not optimize prompts and go with the format provided by the respective model authors or the most intuitive format if none is provided. For each task not designed for evaluating instruction-tuned Code LLMs, we define an instruction. The instruction is to ensure that models behave correctly and that their outputs can be parsed effortlessly.

\begin{figure}[htbp]
\hrulefill

Question: \{context\}\\Is there a defect in the Code, and respond to YES or NO.\\\\Answer: 

\hrulefill
\caption{Prompt for Devign.}
\end{figure}
\FloatBarrier

\begin{figure}[htbp]
\hrulefill

Question: Code 1:     \{context\_1\}\\.\\Code 2:     \{context\_2\}\\Is there a clone relation between the Code1 and Code2, and respond to YES or NO.\\\\Answer: 

\hrulefill
\caption{Prompt for BigCloneBench.}
\end{figure}
\FloatBarrier

\begin{figure}[htbp]
\hrulefill

Question: \{instruction\}\\\{context\}\\\\Answer:\\\{function\_start\}

\hrulefill
\caption{Prompt for HumanEvalPack.}
\end{figure}
\FloatBarrier

\begin{figure}[htbp]
\hrulefill

Question: Create a Python script for this problem.\\\\Answer: \{function\_start\}

\hrulefill
\caption{Prompt for Code Completion on ReCode.}
\end{figure}
\FloatBarrier

\begin{figure}[htbp]
\hrulefill

Question: Create a script for this problem.\\\\Answer: \{function\_start\}

\hrulefill
\caption{Prompt for Asleep At The Keyboard.}
\label{fig:app_devign}
\end{figure}
\FloatBarrier

\section{Timeline}
\paragraph{Sep/2023} Experiment Design; Model Training; Model Evaluation.
\paragraph{Oct/2023} Model Training;  Evaluation Discussion; Model Evaluation.
\paragraph{Nov/2023} Model Evaluation; Result Discussion; Paper Writing.
\paragraph{Dec/2023} Paper Finalization; Codebase Construction.

\end{document}